\documentclass[10pt,twocolumn,letterpaper]{article}

\usepackage{cvpr}              %

\usepackage[dvipsnames]{xcolor}

\newif\ifdrafting
\draftingtrue %
\ifdrafting
   \newcommand{\chen}[1]{\textcolor{cyan}{{[CS: #1]}}}
   \newcommand{\daniel}[1]{\textcolor{red}{DG: #1}}
   \newcommand{\ds}[1]{\textcolor{red}{[ds: #1]}}
   
   \newcommand{\cih}[1]{{\color{green}{[CIH: #1]}}}
   \newcommand{\fc}[1]{\textcolor{blue}{{[FC: #1]}}}
   \newcommand{\ao}[1]{\textcolor{purple}{{[AO: #1]}}}
   \newcommand{\jh}[1]{\textcolor{cyan}{{[JH: #1]}}}
   
\else
   \newcommand{\chen}[1]{}    
   \newcommand{\daniel}[1]{}    
   \newcommand{\ds}[1]{}
   \newcommand{\cih}[1]{}
   \newcommand{\fc}[1]{}
   \newcommand{\ao}[1]{}
   \newcommand{\jh}[1]{}

\fi

\newcommand{\supparxiv}[2]{#2}

\newcommand{\ignore}[1]{}

\newcommand{\ci}[1]{\scriptsize{\textcolor{gray}{~($\pm #1$)}}}

\newcommand{\aftertab}{\vspace{-1.5em}}
\newcommand{\afterfig}{\vspace{-1em}}

\definecolor{cvprblue}{rgb}{0.21,0.49,0.74}
\usepackage[pagebackref,breaklinks,colorlinks,citecolor=cvprblue]{hyperref}
\usepackage{multirow}
\usepackage{xcolor}
\usepackage{makecell}

\usepackage[symbol]{footmisc}

\newcommand{\mypar}[1]{\vspace{2mm}\noindent\textbf{#1}}

\newcommand{\supp}{\emph{sup.~mat.}}

\def\paperID{483} %
\def\confName{CVPR}
\def\confYear{2025}

\title{Motion Prompting: Controlling Video Generation with Motion Trajectories}

\author{
Daniel Geng$^{1,2,*}$ \hspace{0.2cm}
Charles Herrmann$^{1,\dagger}$ \hspace{0.2cm}
Junhwa Hur$^{1}$ \hspace{0.2cm}
Forrester Cole$^{1}$ \hspace{0.2cm}
Serena Zhang$^{1}$ \\
Tobias Pfaff$^{1}$ \hspace{0.2cm} 
Tatiana Lopez-Guevara$^{1}$ \hspace{0.2cm}
Carl Doersch$^{1}$ \hspace{0.2cm}  %
Yusuf Aytar$^{1}$ \hspace{0.2cm}
Michael Rubinstein$^{1}$ \\
Chen Sun$^{1,3}$ \hspace{0.2cm}
Oliver Wang$^{1}$ \hspace{0.2cm}
Andrew Owens$^{2}$ \hspace{0.2cm}
Deqing Sun$^{1}$
\vspace{0.2cm} \\
$^1$Google DeepMind \hspace{0.8cm}
$^2$University of Michigan \hspace{0.8cm}
$^3$Brown University
\vspace{0.2cm} \\
\url{https://motion-prompting.github.io/}
}

\begin{document}
\maketitle
\begin{abstract}

Motion control is crucial for generating expressive and compelling video content; however, most existing video generation models rely mainly on text prompts for control, which struggle to capture the nuances of dynamic actions and temporal compositions. To this end, we train a video generation model conditioned on spatio-temporally sparse \emph{or} dense motion trajectories.
In contrast to prior motion conditioning work, this flexible representation can encode any number of trajectories, object-specific or global scene motion, and temporally sparse motion; due to its flexibility we refer to this conditioning as \emph{motion prompts}. While users may directly specify sparse trajectories, we also show how to translate high-level user requests into detailed, semi-dense motion prompts, a process we term \emph{motion prompt expansion}. 
We demonstrate the versatility of our approach through various applications, including camera and object motion control, ``interacting'' with an image, motion transfer, and image editing.  Our results showcase emergent behaviors, such as realistic physics, suggesting the potential of motion prompts for probing video models and interacting with future generative world models.  Finally, we evaluate quantitatively, conduct a human study, and demonstrate strong performance.

\footnotetext{$^*$Work done as intern, $^\dagger$Project lead}
\end{abstract}    
\section{Introduction}
\label{sec:intro}

\begin{figure*}[t]
    \centering
    \includegraphics[width=\linewidth]{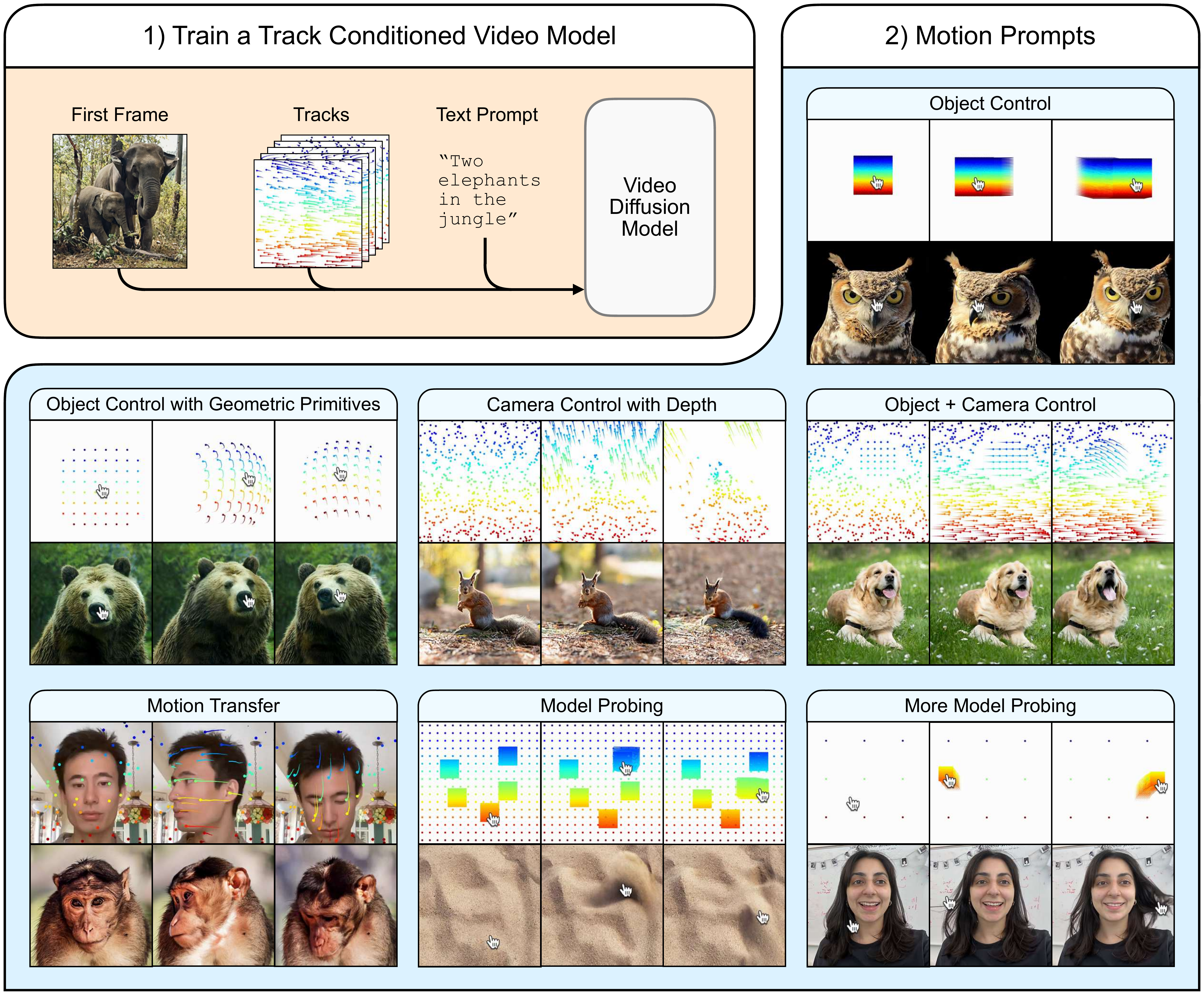}
    \caption{{\bf Motion Prompting.} 1) We train a general-purpose track-conditioned ControlNet adapter on top of a video diffusion model. 2) To use this model, we design {\it motion prompts} from user inputs, and show a  variety of capabilities from this single trained model, such as object control, camera control, simultaneous object and camera control, motion transfer, and model probing. We visualize the motion prompt tracks and corresponding frames from the generated videos underneath. The tracks are colored only for the purpose of visualization, with trails denoting the direction and magnitude of motion. Additionally, some of our motion prompts are derived from user mouse motions, for which we visualize the mouse locations.
    \textcolor[RGB]{121, 121, 255}{\textbf{We highly encourage the reader to view video results \supparxiv{in the \supp{}}{on our \href{https://motion-prompting.github.io/}{webpage} .} }} }
    
\label{fig:teaser}
\afterfig
\end{figure*}

In video generation, motion is paramount. It can elevate a video from the uncanny valley to realistic or from amateur to professional. Motion guides attention, enhances storytelling, and defines visual style. Skilled filmmakers, like Kubrick and Kurosawa, masterfully use motion to create captivating, immersive experiences.
Achieving realistic and expressive motion, coupled with granular control, is essential for generating compelling video. While text remains the main control signal for generation, its limitations become apparent when focusing on motion.
Although effective for describing static scenes in images or high level actions, text struggles to convey the subtleties of motion:
\eg, a prompt like ``a bear quickly turns its head'' could be interpreted in countless ways. How quick is ``quickly''? What is the exact trajectory? Should it accelerate? 
Even detailed descriptions fail to capture nuances like ease-in-ease-out timing or synchronized movements.
These nuances are often better conveyed through the motion itself.

Motivated by this, we explore motion as a powerful, complementary control scheme to text. Our first observation is that in order to fully harness the expressiveness of motion, we require a representation that can encode \emph{any} type of motion.
To this end, we identify spatio-temporally sparse \emph{or} dense motion trajectories~\cite{sand2008particle, harley2022particle} as an ideal candidate. Motion trajectories, a.k.a.~particle video or point tracks, track the movement and visibility of a set of points throughout a video, offering a highly expressive encoding of motion. This representation can capture the trajectories of any number of points, represent object-specific or global scene motion, and even handle temporally sparse motion constraints.
Furthermore, recent advances in point track estimation have yielded robust and efficient algorithms \cite{doersch2023tapir, doersch2024bootstap,karaev23cotracker,karaev2024cotracker3} that are capable of processing diverse real-world videos to generate constraints for training. 
Given the comprehensive and flexible nature of this motion representation, akin to text, we designate our motion conditioning as \emph{motion prompts}. 
We then train a motion trajectory ControlNet~\cite{zhang2023adding} on top of a pre-trained video diffusion model~\cite{bar2024lumiere} to accept the motion prompt conditioning.

While these motion prompts can define any type of video motion, what is less clear is how a user would generate them in practice. 
Sparse trajectories, which give the rough direction of a few pixels or patches, may be easy to specify with mouse drags, but do not sufficiently constrain the generation process and fall short with respect to fine-grained control. Conversely, dense trajectories, though offering precise control, are impractical to design manually. 
To address this, our second observation is that we can often translate high level user requests (\eg, ``move the camera around the $xz$ plane'', ``rotate the head of the cat'') into detailed motion trajectories through computer vision signals. We denote this process as \emph{motion prompt expansion} due to its similarities to prompt expansion~\cite{datta2023prompt} or rewriting~\cite{betker2023improving} for text in image generation. This method is intended to bridge the gap between user goals and our  motion representation.

We identify several instances where motion prompt expansion can be an effective tool including (\cref{fig:teaser}): converting user mouse drags into semi-dense motion trajectories allowing users to ``interact'' with an image by manipulating hair or sand (\cref{sec:interacting});  simultaneously specifying camera and object motion (\cref{sec:composition}); performing motion transfer where motion from a given video is applied to a different first frame (\cref{sec:transfer}); and performing drag-based image editing (\cref{sec:interacting}). While these results are not yet real time or causal, they strongly hint at how users may interact with generative world models, and allows us to probe the video prior of the generator to understand the aspects of physics and general world knowledge it has learned.

Finally we present quantitative results and human studies against baselines, indicating that our model performs well. We also present ablations to validate our design choices and give insight. In summary, our contributions are:

\begin{itemize}
    \item We focus on motion as a conditioning signal and identify spatio-temporally sparse \emph{or} dense motion tracks as a flexible motion representation that can accomplish many aspects of motion control. We train a ControlNet to accept these \emph{motion prompts} as conditioning.
    \item We propose \emph{motion prompt expansion}, a process which takes simple user input and produces more complex motion tracks, which allow for more fine-grained control. 
    \item We then apply our approach to a wide range of tasks, such as object control, camera control, motion transfer, or drag-based image editing. 
    \item We also show emergent behavior, such as physics, which suggests that these motion prompts may be used to probe video models or interact with future world models.
    \item We evaluate our method against baselines with quantitative metrics and a human study, showing that our model performs well compared to baselines.
\end{itemize}

\section{Related Work}
\label{sec:related_work}

\mypar{Video Diffusion Models.}
Diffusion models~\cite{sohldickstein2015diffusion,ho2020denoising,song2020denoising} have demonstrated amazing capabilities for video generation, conditioned on natural language~\cite{ho2022video,ho2022imagen,guo2023animatediff,girdhar2023emu,bar2024lumiere} or by ``animating'' static images into videos~\cite{singer2022make,blattmann2023stable,xing2024dynamicrafter}.
Beyond content creation~\cite{polyak2024movie}, they can be seen as a path to the ambitious goal of creating world simulators~\cite{videoworldsimulators2024}, showing preliminary success in visual planning for embodied agents~\cite{du2024learning,escontrela2024video,yang2024video}.
Meanwhile, whether the video prior captures sufficient understanding of the physical world is still under debate~\cite{kang2024far}, and explicit integration of physics rules appears to be necessary~\cite{yuan2023physdiff,liu2024physgen,yu2024lucidsim}. 
Our motion prompting technique, applicable to any video diffusion model, not only offers a more flexible and accurate interface to specify motion patterns for video generation, but also serves as a framework to probe %
a trained generative model for their 3D or physics understanding.

\mypar{Motion-conditioned Video Generation.}
A pre-trained text-to-video model can be adapted to follow new motion patterns or additional motion conditioning signals.
Low-rank adaptation (LoRA)~\cite{hu2021lora}, a generic technique for parameter fine-tuning, can be utilized for few-shot motion customization~\cite{ren2024customize,zhao2025motiondirector}.
DreamBooth~\cite{ruiz2022dreambooth}, originally for personalized image generation, can also be applied to video generation~\cite{wu2024motionbooth} with motion control.

Early work proposes video control through sparse motions~\cite{hao2018controllable,ardino2021click}. More recent work explores similar ideas with more powerful models. %
The approaches vary in their design choices but often require certain complicated engineering techniques for stable training and better convergence.
Tora~\cite{zhang2024tora}, MotionCtrl~\cite{wang2024motionctrl}, DragNUWA~\cite{yin2023dragnuwa}, Image Conductor~\cite{li2024image}, and MCDiff~\cite{chen2023mcdiff} adopt two-stage (\eg, finetuning with first dense and then sparse trajectories, or training adapters sequentially), specialized losses~\cite{li2024image,namekata2024sg}, architectures~\cite{xiao2024trajectory,feng2024i2vcontrol}, or multi-stage fine-tuning for multiple modules~\cite{2023videocomposer,chen2023mcdiff,shi2024motion}.
MOFA-Video~\cite{niu2024mofa} requires separate adapters for different motion types, TrackGo~\cite{zhou2024trackgo} uses custom losses and layers, while other works~\cite{yin2023dragnuwa,wang2024motionctrl,niu2024mofa,li2024image,zhang2024tora} engineer data filtering pipelines.
In contrast, we find that a simpler training recipe yields high quality results. Our model is trained in a single stage, with uniformly sampled dense trajectories, and without any specialized engineering efforts. %
Yet it handles a wide range of tasks and motions, generalizing to both sparse and dense trajectories during inference.

Other approaches use entity-centric control signals such as bounding boxes~\cite{wu2024motionbooth,wang2024boximator}, segmentation masks~\cite{wu2024draganything, dai2023finegrained}, human pose~\cite{hu2023animateanyone,xu2023magicanimate}, or camera pose~\cite{he2024cameractrl,watson2024controlling}.
Zero-shot motion adaptation approaches (\eg,~SG-I2V~\cite{namekata2024sg}, Trailblazer~\cite{ma2023trailblazer}, FreeTraj~\cite{qiu2024freetraj}, and Peekaboo~\cite{jain2024peekaboo}) adopt a similar strategy, guiding the video generation based on the motion of entity-centric masks and thus avoiding training or fine-tuning video models.
Our motion prompts offer a more flexible interface to control motion generation at various granularity. Unlike the test-time approaches which explicitly control the diffusion feature maps, our framework naturally balances the strength of controlling signals and that of the encoded video priors.

\mypar{Motion Representations.} 
As our goal is to condition a video generation model on motion of any type, it is crucial to choose a suitable motion representation. 
The most common representation is optical flow~\cite{horn1981determining,lucas1981iterative,brox2010large,dosovitskiy2015flownet,sun2018pwc,teed2020raft}.
While flow can be chained over time, errors can accumulate.
The lack of occlusion handling also makes it unsuitable for our needs, which we find necessary for good camera control (\cref{sec:depth}).
In contrast, long-range feature matching~\cite{jabri2020space,bian2022learning,shrivastava2024selfsupervised,jiang2021cotr} or point trajectories~\cite{harley2022particle,zheng2023point,karaev23cotracker,karaev2024cotracker3,doersch2023tapir,doersch2024bootstap} is a well-suited representation for our application. It can handle occlusions and allows for both sparse and dense tracking over any arbitrary temporal durations.

\section{Method}

Our video generation method takes as input a single frame, a text prompt, and a motion prompt in the form of point tracks---which we explain how to create in Sec.~\ref{sec:motionprompting}. Full implementation details can be found in \cref{apdx:implementation}. %

\subsection{Motion Prompts}

To fully harness the expressiveness of motion, we need to be able to represent any type of motion. 
To this end, we use point trajectories for our motion prompts, which can encode both spatially (and temporally) sparse and dense motions, motion on a single object or of an entire scene, and even occlusions via a visibility flag.
Using this representation enables a broad range of capabilities such as object control (\cref{sec:interacting}), camera control (\cref{sec:depth}), both simultaneously (\cref{sec:composition}), motion transfer (\cref{sec:transfer}), and drag-based image editing (\cref{sec:interacting}) under a unified model.

Formally, we denote a set of $N$ point trajectories of length $T$ by $\mathbf{p} \in \mathbb{R}^{N \times T \times 2}$, where the 2D coordinate of the $n^{th}$ track at the $t^{th}$ timestep is $\mathbf{p}[n, t] = (x_t^n, y_t^n)$.  In addition, we denote the visibility of the tracks as $\mathbf{v} \in \mathbb{R}^{N \times T}$, an array of 1's and 0's where $0$ indicates an off-screen or occluded track, and $1$ indicates a visible track.

\subsection{Architecture}

\begin{figure}[t]
    \centering
    \includegraphics[width=\linewidth]{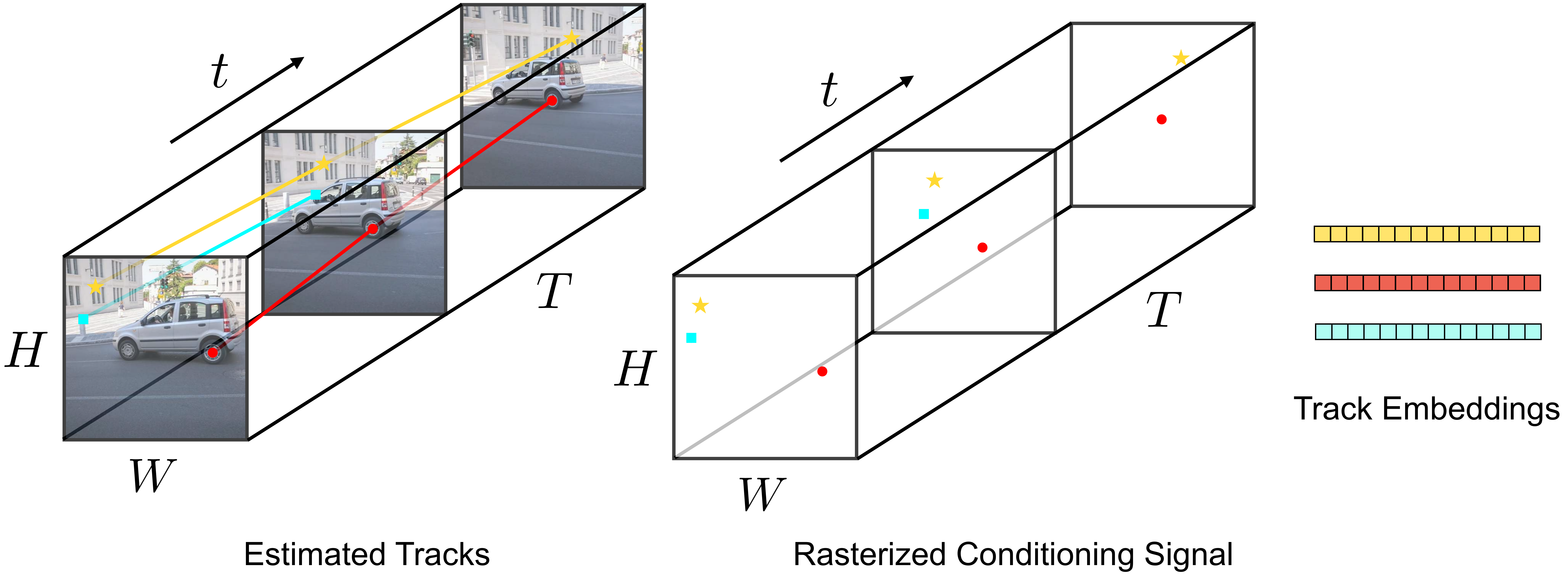}
    \caption{{\bf Conditioning Tracks.} During training, we take estimated tracks from a video (left) and encode them into a $T \!\times\! H \!\times\! W \!\times\! C $ dimensional space-time volume (middle). Each track has a unique embedding (right), written to every location the track visits and is visible at. All other locations are set to zeros. This strategy can encode any number and configuration of tracks. 
    }
\label{fig:conditioning}
\afterfig
\end{figure}

We build our model on top of Lumiere, a pre-trained video diffusion model~\cite{bar2024lumiere} which has been trained to generate 5 seconds of video at 16 fps given text and first frame conditioning.
In order to train in track conditioning, we use a ControlNet~\cite{zhang2023adding} which requires encoding tracks in a spatial-temporal volume, $\mathbf{c} \in \mathbb{R}^{T \times H \times W \times C}$, where $T$ is the number of frames, $H$ and $W$ are the height and width of the generated video, and $C$ is the channel dimension. To do this, we associate with each track, $\mathbf{p}[n, :]$, a unique and random embedding vector $\phi^n \in \mathbb{R}^C$. Then, for each space-time location a track visits, and is visible at, we simply place the embedding $\phi^n \in \mathbb{R}^C$ in that location. All other values in the conditioning signal are set to 0.
\cref{fig:conditioning} illustrates this process. In other words, we zero-initialize $\mathbf{c}$ and set
\begin{equation}
    \mathbf{c}[t, x_t^n, y_t^n] = \mathbf{v}[n,t] \phi_n
\end{equation} 
for all tracks at each timestep $t$, where multiplying by the visibility $\mathbf{v}[n, t]$ zeros out the embedding if the track is not visible at that location and time. We quantize $x_t^n$ and $y_t^n$ to the nearest integer for simplicity. When multiple tracks pass through the same space-time location, we add the embeddings together. 
The track embeddings $\phi^n$ are {\it randomly} drawn from a fixed pool, and act simply as a unique identifier for each track. %
For completely dense tracks, this representation is equivalent to starting with a dense grid of embeddings and forward warping, similar to~\cite{seo2024genwarp}.

\subsection{Data}

To train our model, we prepare a video dataset paired with tracks. We run BootsTAP~\cite{doersch2024bootstap}, an off-the-shelf point tracking method, on an internal dataset consisting of 2.2M videos resized to $128 \times 128$, the output size of our base model. 
We extract tracks densely, resulting in 16,384 tracks per video as well as predicted occlusions, which we can sample from during training. We do not filter the videos in any way, with the hypothesis being that training on diverse motions will result in a more powerful and flexible model.

\subsection{Training}
\label{sec:training}

Training follows ControlNet~\cite{zhang2023adding}, where the conditioning signal is given to a trainable copy of the base model's encoder and the standard diffusion loss is optimized. For every video, we sample a random number of tracks from a uniform distribution and construct the conditioning signal as explained above. More details can be found in \cref{apdx:implementation}.

We observe various phenomena during training. For one, we find that the loss is not correlated with the performance of the model at following tracks. Also, similar to~\cite{zhang2023adding}, we observe a ``sudden convergence phenomena" in which the model goes from completely ignoring the conditioning signal to fully trained in a short number of training steps. More details can be found in \cref{apdx:training}.

Finally, we observe that our model exhibits fairly strong generalization in multiple directions. For example, while our model is trained on randomly sampled tracks, resulting in spatially uniformly distributed tracks during training, the model can generalize to spatially localized track conditioning (\cref{fig:sparse,fig:primitives}). In addition, while our model is trained for specific numbers of tracks, it generalizes surprisingly well to more (\cref{fig:depth}) or fewer number of tracks (\cref{fig:sparse,fig:primitives,fig:drag_editing}). Finally, we find that our model generalizes to tracks that don't necessarily start from the first frame, despite only being trained on these tracks (\cref{fig:sparse}b). We hypothesize this generalization is due to a combination of inductive biases from the convolutions in the network and the fact that we train the model on a large variety of trajectories.

\section{Motion Prompts}
\label{sec:motionprompting}

In this section, we discuss different types of effects achievable through motion prompts and prompt expansion. In particular, we identify and demonstrate several different types of expansion, as shown in \cref{fig:teaser}. Text prompts and other parameters for each video may be found in \cref{tbl:figure_details}. \textbf{We strongly encourage readers to view generated videos \supparxiv{in the \supp{}}{on our \href{https://motion-prompting.github.io/}{webpage}.}}

\subsection{``Interacting'' with an Image} 
\label{sec:interacting}

\begin{figure*}[t]
    \centering
    \includegraphics[width=\linewidth]{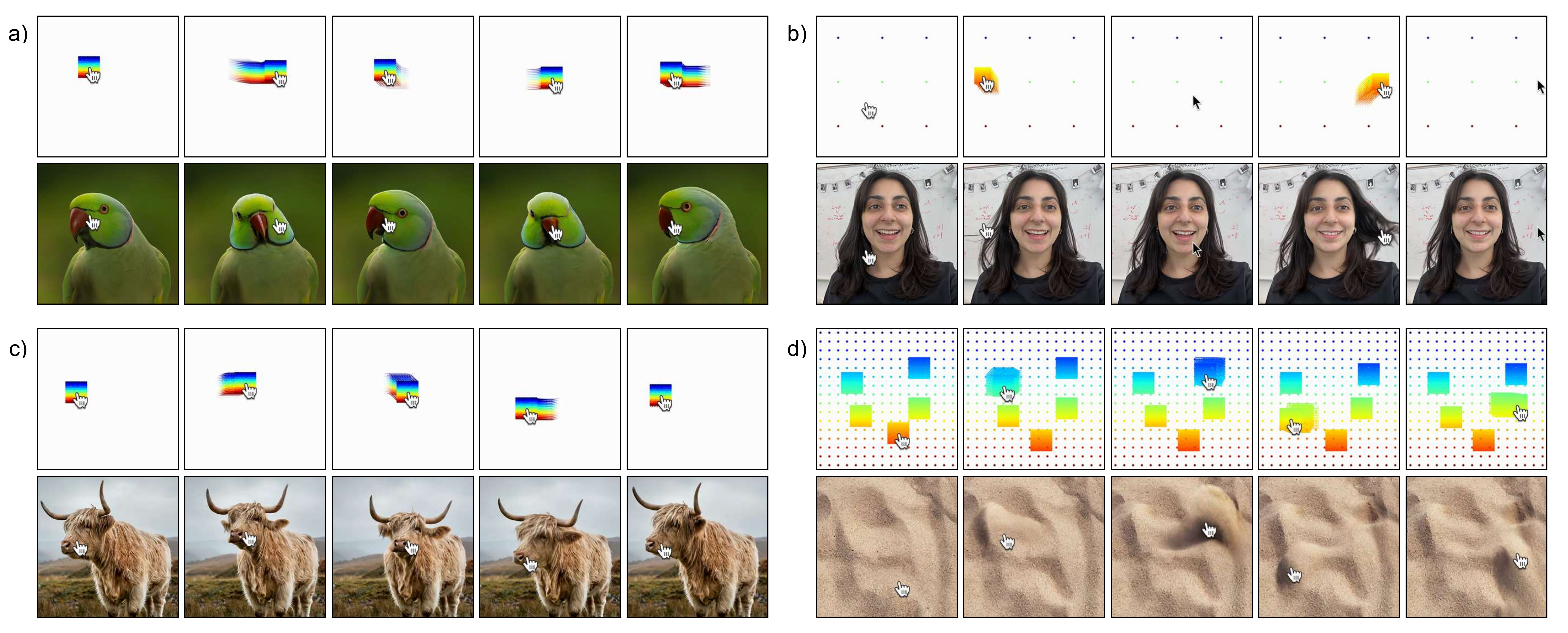}\vspace{-3mm}
    \caption{{\bf ``Interacting" with an Image.} We translate a simple user input, mouse motions and drags, and expand it into a more complex motion prompt which helps to achieve the user's intention. The mouse trajectories are visualized as a hand when dragging, and as a black cursor otherwise. A grid of tracks centered on the cursor are created when the mouse is dragged, as shown in the top row. Frames from the generated video are shown in the bottom row. Prompting our model in this way, we can (a) move the head of a parrot or (c) a cow (b) play with hair or (d) ``interact" with an image of sand. We can also keep the background still by specifying static tracks, as in (b) or (d). \textbf{Note these samples are not generated in real-time and are not temporally causal. More examples can be found \supparxiv{in the \supp{}}{on our \href{https://motion-prompting.github.io/}{webpage}.}}
    }
\label{fig:sparse}
\vspace{2mm} %
\afterfig
\end{figure*}
\begin{figure}[t]
    \centering
    \includegraphics[width=\linewidth]{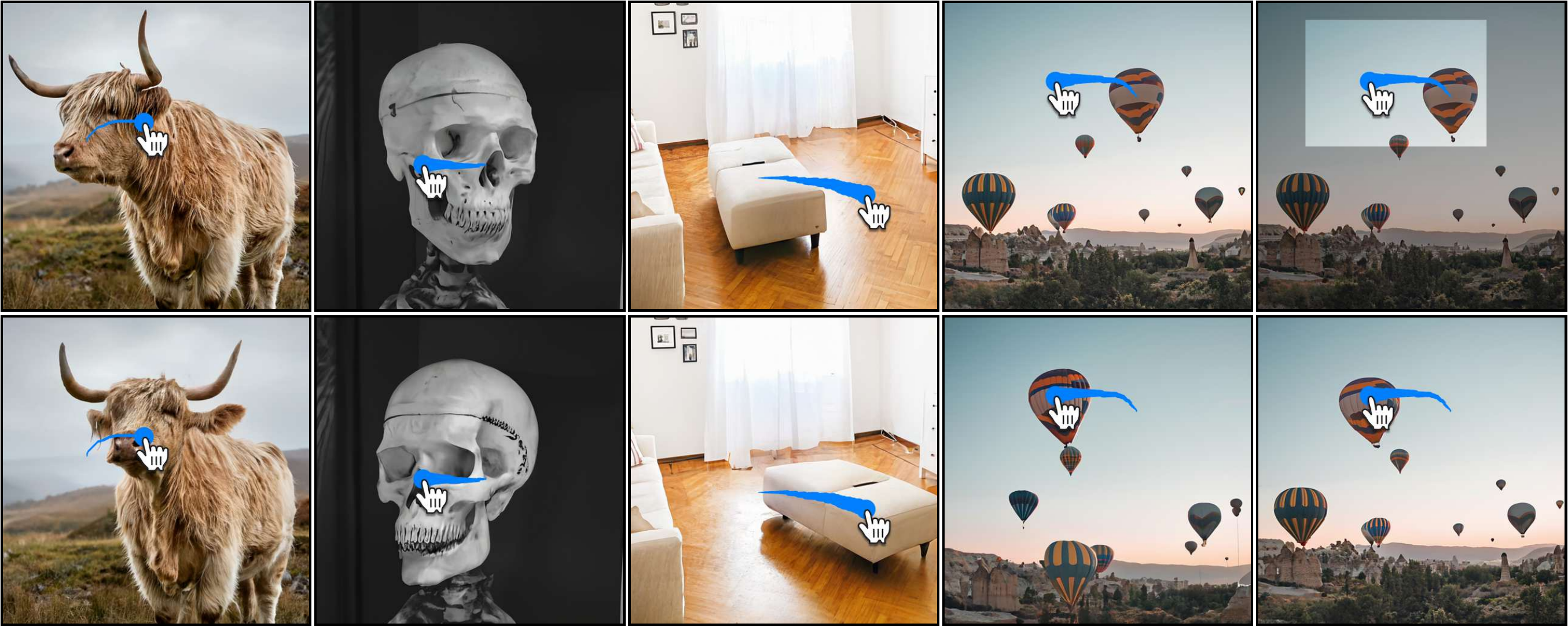}
    \caption{{\bf Drag-Based Image Editing.} We show the input images in the first row, and resulting drag-based edits in the bottom row, with the drag visualized in both rows. In addition, in the final example we show how we can keep areas of the images static.}
\label{fig:drag_editing}
\afterfig
\end{figure}

Our model enables the ability to ``interact" with images. To do this, we build a GUI that displays a still image and records mouse drags from a user.
This recording is then converted to tracks, as described below, and is fed to the model along with the initial frame and text.
More information about the GUI can be found in \cref{apdx:implementation}.
For simple mouse motions, where the mouse is constantly dragged, this approach is similar to prior work on sparse trajectory conditioned video generation~\cite{wang2024motionctrl,yin2023dragnuwa,li2024image,wu2024draganything,niu2024mofa,2023videocomposer,shi2024motion,zhang2024tora,zhou2024trackgo,namekata2024sg}.
However, because our model generalizes to partial tracks, we can also handle multiple mouse drags in different locations at different times, resulting in natural user control as in \cref{fig:sparse}b and \cref{fig:sparse}d.
Please note that while we record mouse inputs in real-time, our method requires sampling from the video diffusion model, which is not real-time -- it takes about 12 minutes to generate an output video. 

To create the motion prompts, we translate mouse drags into a grid of point tracks as shown in \cref{fig:sparse}. 
The density and size of this grid can be chosen by the user, similar to the Gaussian blurring of tracks in~\cite{yin2023dragnuwa,wang2024motionctrl,wu2024draganything,li2024image} to specify the spatial extent of the motion.
However, note that in our approach this step is done only at inference time, and not at train time.
Additionally, a user may choose to place a grid of static tracks down to keep the background still, as in \cref{fig:sparse}b and \cref{fig:sparse}d, or have tracks persist after a mouse drag as in \cref{fig:sparse}d.

\mypar{Emergent Phenomena.} We find that these ``interaction" motion prompts can result in straightforward motions, such as turning the head of a parrot in \cref{fig:sparse}a. But interestingly, we also observe more complex dynamics: \eg, in \cref{fig:sparse}b, where the tracks toss the hair of the subject, or in \cref{fig:sparse}d where the sand is swept around. In these examples, we are essentially probing the video prior learned by the model, and by doing so are able to visualize the physics and general world understanding that the model has learned. Furthermore, because our model supports temporally sparse track conditioning, we can effectively do prediction. That is, we can query our model with a motion for a short duration, and then let the model predict the future, allowing us to answer questions such as {\it "how will the hair behave if I pull on it this way or that?"} as in \cref{fig:sparse}b.

\begin{figure*}[t]
    \centering
    \includegraphics[width=\linewidth]{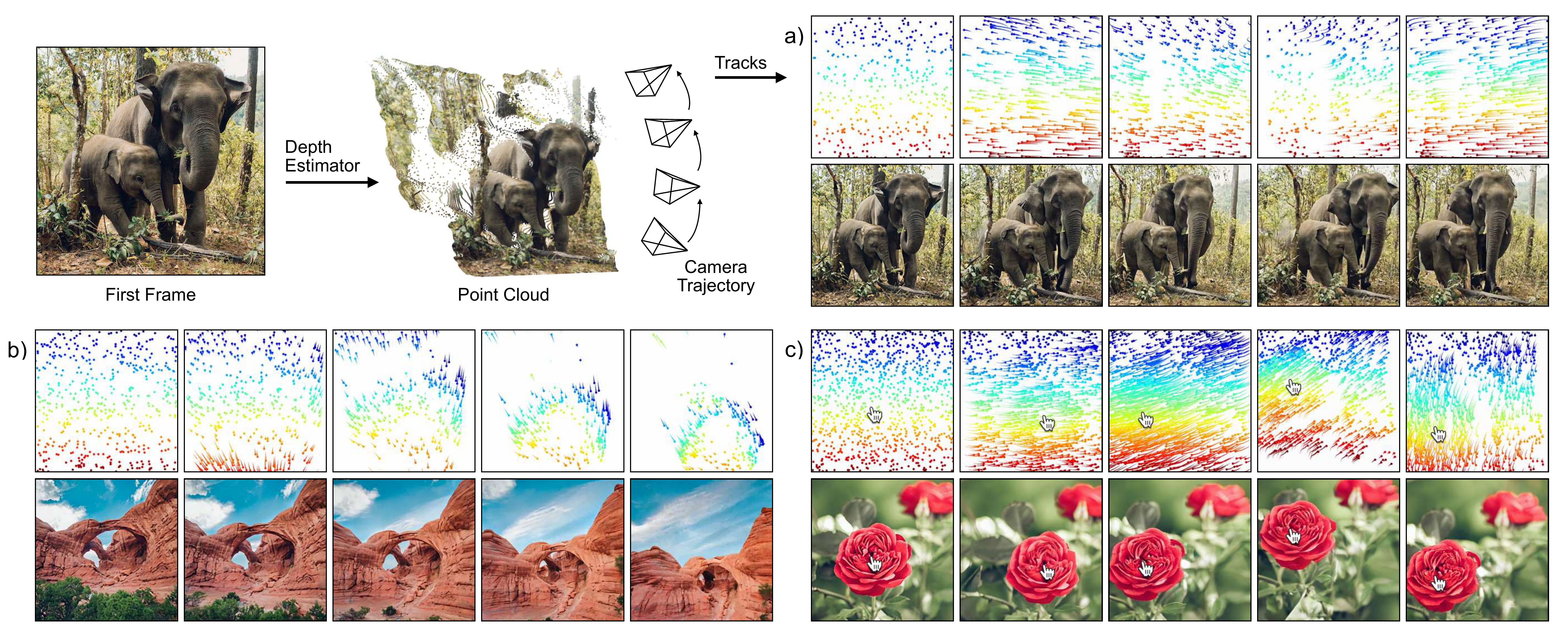}\vspace{-3.5mm}
    \caption{{\bf Camera Control with Depth.} We can construct motion prompts for camera control by specifying a camera trajectory and computing a point cloud with an off-the-shelf monocular depth estimator. We then project the points onto the sequence of cameras, which results in the shown point trajectories. We can also convert user mouse input into camera trajectories, as in example (c).}
\label{fig:depth}
\afterfig
\end{figure*}
\begin{figure}[t]
    \centering
    \includegraphics[width=0.95\linewidth]{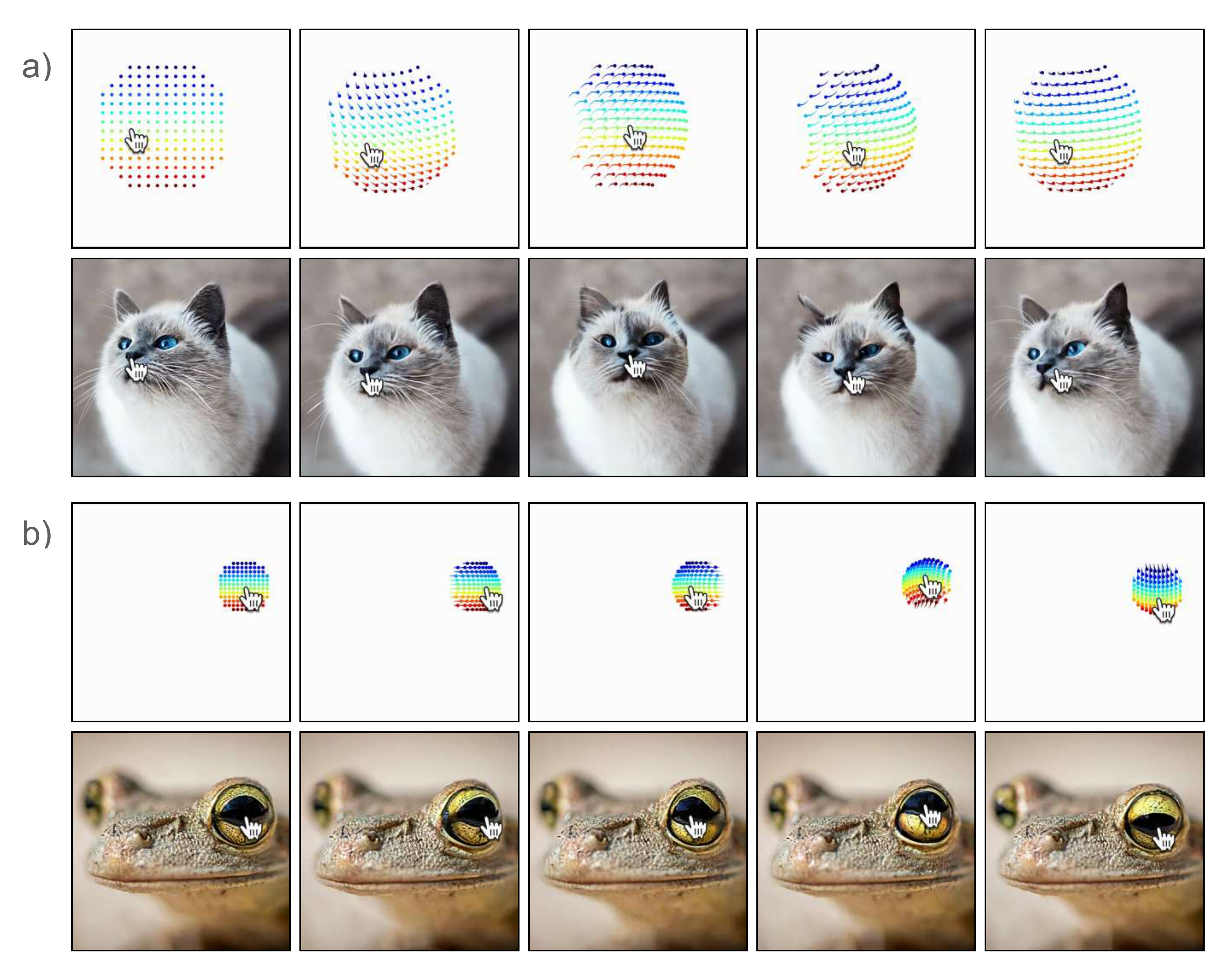}\vspace{-3.5mm}
    \caption{{\bf Object Control with Primitives.} By defining geometric primitives (\eg, a sphere) manipulated by a user with a mouse, we can obtain tracks exerting more fine-grain control over objects (\eg, rotations), which cannot be specified with a single track.}
\label{fig:primitives}
\afterfig
\end{figure}

\mypar{Drag-Based Image Editing.} A natural application of this ``interaction" ability is drag-based image editing~\cite{pan2023drag,shi2023dragdiffusion,mou2023dragondiffusion,geng2024motion,alzayer2024magic,rotstein2024pathways}. This task involves taking user supplied ``drags" and editing an image such that objects follow these drags.
We shows qualitative results in \cref{fig:drag_editing}.

\subsection{Object Control with Primitives}
\label{sec:primitives}

We can also reinterpret mouse motions as manipulating a proxy geometric primitive, such as a sphere. By placing these tracks over an object that can be roughly approximated by the primitive, we can effect more fine-grained control over the object than with sparse mouse tracks alone. For example, in \cref{fig:primitives}, we place a sphere over the head of a cat and the eye of a frog to precisely rotate these objects to different positions, and in \cref{fig:teaser} we animate a bear. In this setting, the user must supply both the mouse motion, and also the location and radius of the sphere to use.
This allows for the user to specify more complex motions than translations, which would be hard to express with a single mouse-drag created trajectory. 
For implementation details, please see \cref{apdx:implementation}. %

\subsection{Camera Control with Depth}
\label{sec:depth}

We can also design motion prompts to achieve camera control with our model. We do this by first running an off-the-shelf monocular depth estimator~\cite{piccinelli2024unidepth} on the input frame to obtain a point cloud of the scene. Then, given a trajectory of camera poses we can re-project the point cloud onto each camera, resulting in 2D tracks for input.
We can further improve quality by running z-buffering to get occlusion flags.

Prompting our model with these motion prompts results in camera control, as shown in \cref{fig:depth}. We can orbit a camera in circles as in \cref{fig:depth}a or have it arc upwards as in \cref{fig:depth}b. In addition, we can combine this camera control with mouse recordings for even greater ease of use. To do so, we record mouse inputs as is done in \cref{sec:interacting}. We then construct a camera trajectory such that a single point in the point cloud follows the mouse trajectory, and that the camera is constrained to a vertical plane, which we show in \cref{fig:depth}c.
For implementation details, please see \cref{apdx:implementation}.

Note that our model is neither trained on nor conditioned on camera poses, as with prior work~\cite{wang2024motionctrl,li2024image,watson2024controlling}. Furthermore, none of our training data includes pose annotations. 
Despite this, we find that our model can achieve compelling camera control. 
This shows that instead of training a video model on specific types of motion, we can train a model on general motions and tease out specific capabilities by using motion prompts.

\subsection{Composing Motions}
\label{sec:composition}

\begin{figure}[t]
    \centering
    \includegraphics[width=0.95\linewidth]{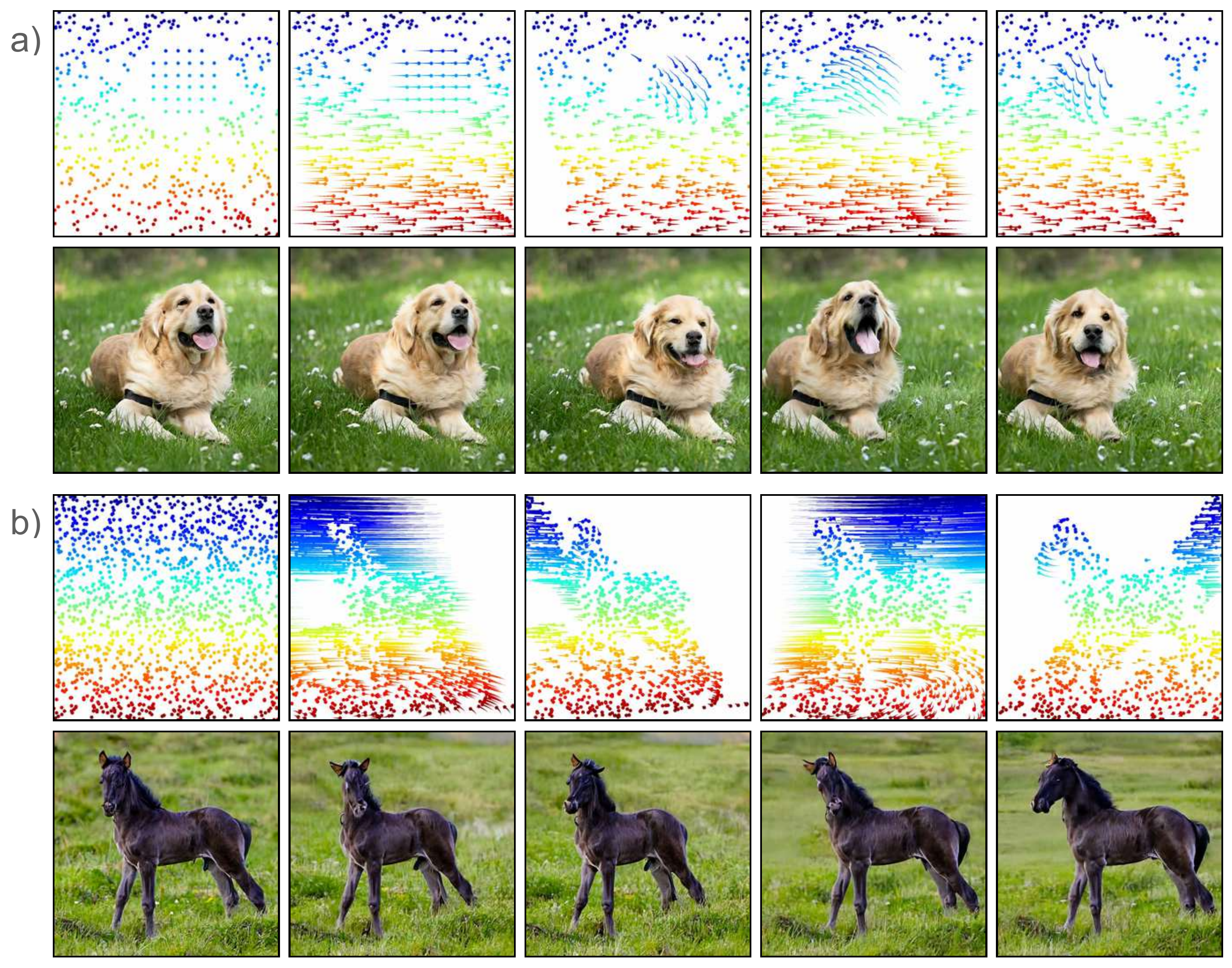}
    \vspace{-2mm}
    \caption{{\bf Compositions of Motion Prompts.} By composing motion prompts together, we can attain simultaneous object and camera control. For example, here we move the dog and horse's head while orbiting the camera from left to right.}
\label{fig:compositions}
    \vspace{-2mm}
\afterfig
\end{figure}

By composing motion prompts together we can combine capabilities. For example, in \cref{fig:compositions}, we add together tracks for object control and camera control, resulting in simultaneous control of both. This is done by converting the object tracks to displacements, and adding these deltas to the camera control tracks. In two dimensions, this composition is an approximation and will fail for extreme camera motion, but we find it works well for small to moderate camera motion. 
Again note, that we do not specifically train for this capability in contrast to prior work~\cite{wang2024motionctrl,li2024image,watson2024controlling}.

\subsection{Motion Transfer}
\label{sec:transfer}

\begin{figure}[t]
    \centering
    \includegraphics[width=0.95\linewidth]{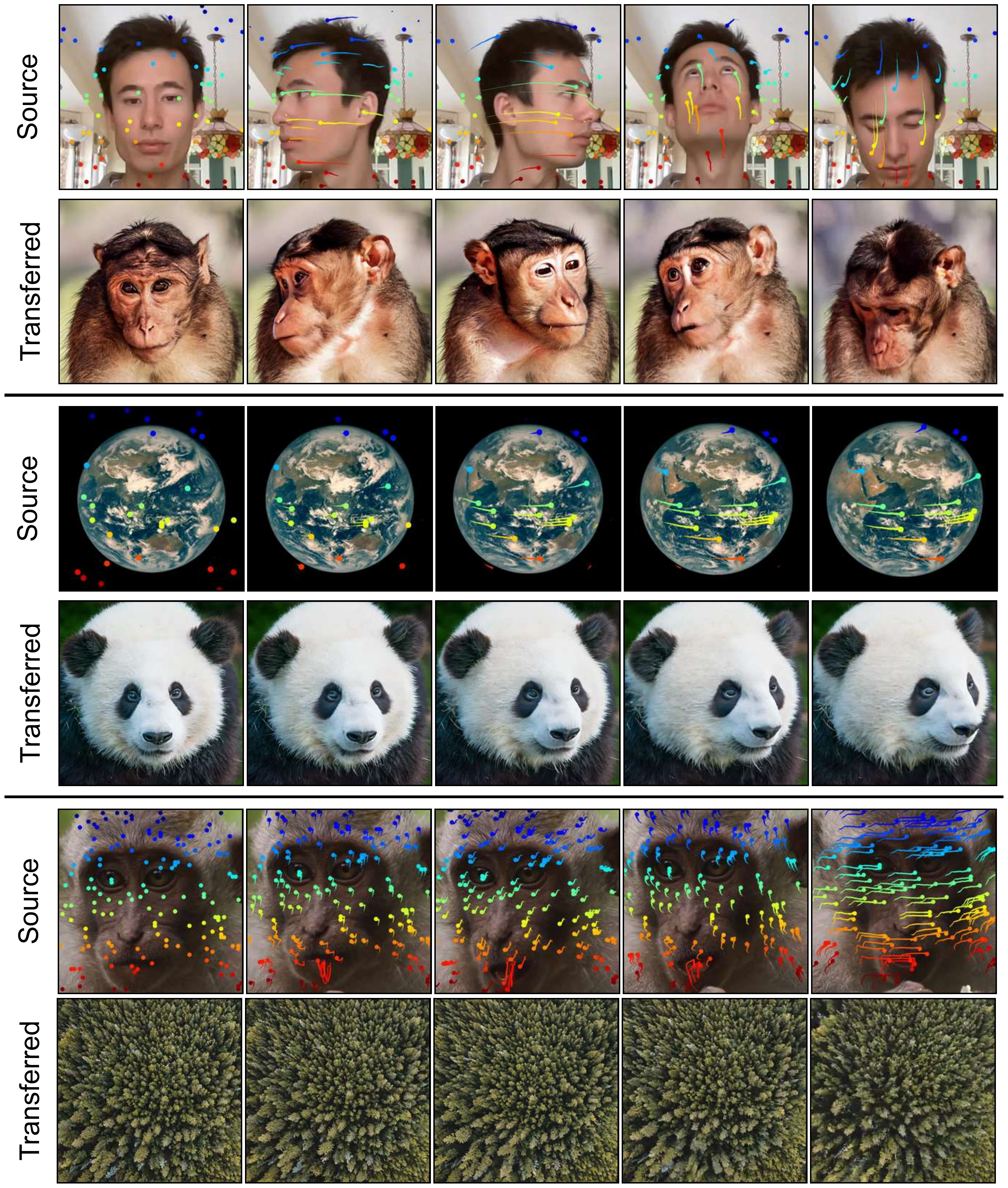}\vspace{-2mm}
    \caption{{\bf Motion Transfer.} By conditioning our model on extracted motion from a source video we can puppeteer a macaque, or even transfer the motion of a {\it monkey chewing} to a photo of trees.  
    \textbf{
    Best viewed as videos \supparxiv{in the \supp{}}{on our \href{https://motion-prompting.github.io/}{webpage}}.} 
    }
\label{fig:motion_transfer}
\afterfig
\end{figure}

Many types of motions may be hard to design a motion prompt for. Given a video with a desired motion, we can perform motion transfer~\cite{2023videocomposer,geng2024motion}, where we extract motion tracks from a source video and apply it to an image. For example, we can extract the motion of a person turning their head and use it to puppeteer a macaque, as in \cref{fig:motion_transfer}. Moreover, we find that our model is surprisingly robust, in that we can apply motions to fairly out-of-domain images. For example, in \cref{fig:motion_transfer} we apply the motion of {\it a monkey chewing} to a bird's eye view image of trees. 
The resulting videos exhibit an interesting effect in which pausing the video on any frame removes the percept of the source video~\cite{reddit2024rickroll}. 
The {\it monkey} can only be perceived when the video is playing, where a Gestalt common-fate effect occurs~\cite{johansson1973visual}.

\begin{figure}[t]
    \centering
    \includegraphics[width=0.95\linewidth]{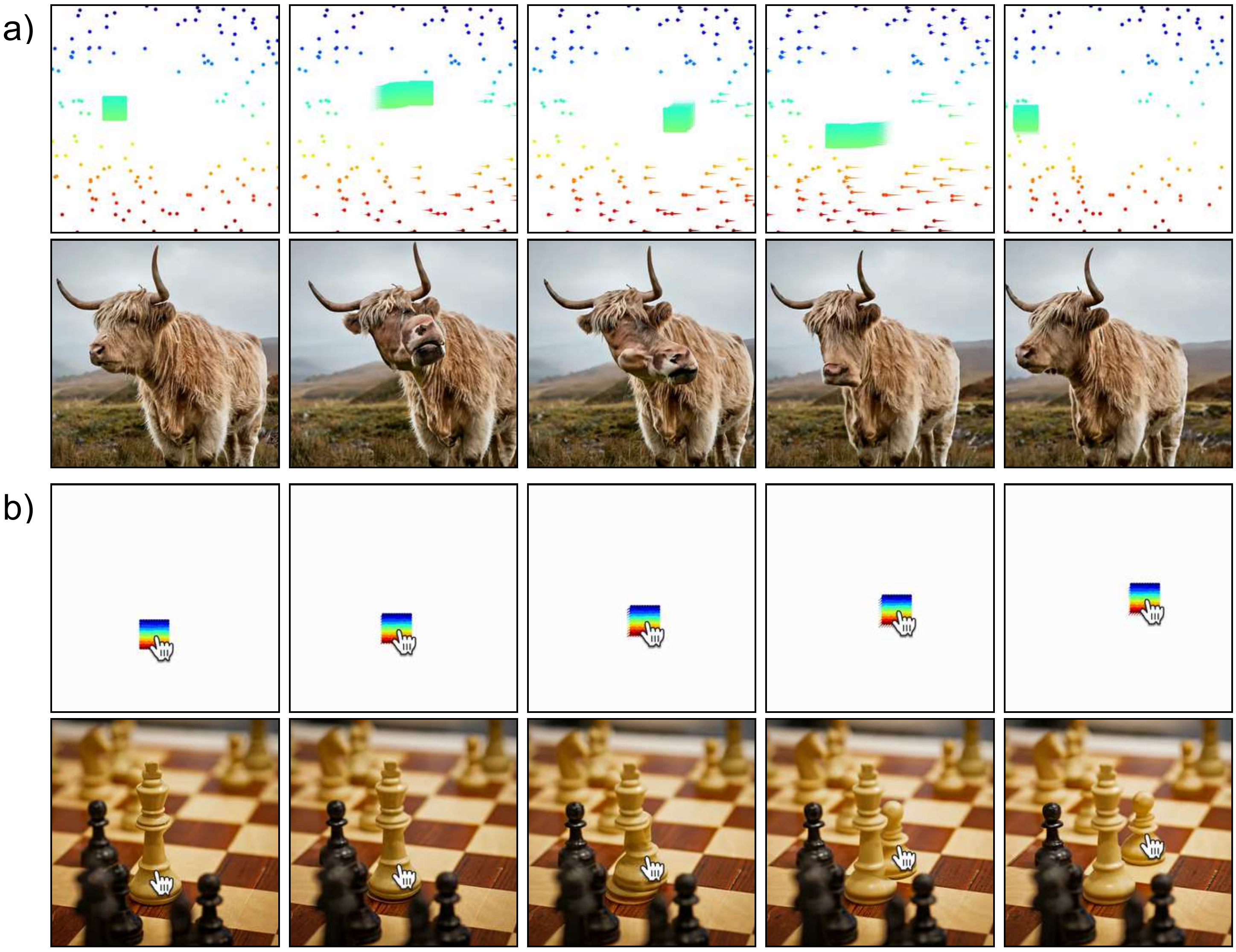}
    \caption{{\bf Probing by Failures.} We can use motion prompts to probe limitations of the underlying model. For example, dragging the chess piece results in the creation of a new piece.}
\label{fig:failures}
\afterfig
\end{figure}

\subsection{Failures, Limitations, and Probing Models}
\label{sec:failures}
We differentiate failures into two broad categories. The first are failures of our motion conditioning or our motion prompting. For example, in \cref{fig:failures}a we show an example in which the cow's head is unnaturally stretched due to the horns being mistakingly ``locked" to the background. 
The second category are failures due to the underlying video model. 
For example, in \cref{fig:failures}b, we drag the chess piece but a new one spontaneously forms, which is a less plausible video given the constraint. 
These types of failures suggest that we might be able to use motion prompts as a way to {\it probe} video models and discover limitations in their learned representations.

\section{Quantitative Results}
\label{sec:quantitative}

In addition to the qualitative examples above, we describe a quantitative benchmark, and evaluate our method against recent baselines. Furthermore, we conduct a human study, and describe ablations in this section.

\subsection{Track-Conditioned Generation Evaluation}
\label{sec:davis_eval}

To evaluate our track-text-and-first frame conditioned video generation, we use the validation split of the DAVIS video dataset~\cite{pont20172017}. We extract first frames and tracks from the dataset and feed this to the models along with an automatically generated text prompt. For exact implementation details, please see \cref{apdx:implementation}. 
To evaluate a range of track densities, we vary the number of conditioning tracks from just a single track to 2048 tracks.

We compare our method with two recent works: ImageConductor~\cite{li2024image}, which finetunes AnimateDiff~\cite{guo2023animatediff} for camera and object motion, and DragAnything~\cite{wu2024draganything}, which is designed to move ``entities" along tracks by finetuning Stable Video Diffusion~\cite{blattmann2023stable}.
To evaluate the appearance of the generated videos we compute \textbf{PSNR}, \textbf{SSIM}~\cite{wang2004image}, \textbf{LPIPS}~\cite{zhang2018perceptual}, and \textbf{FVD}~\cite{unterthiner2019fvd} between the generated videos and ground truth videos. 
To evaluate how well the generated video matches the motion conditioning, we use end-point error (\textbf{EPE}) which is computed as the L2 distance between the conditioning tracks and tracks estimated from the generated videos.

\begin{table}[t]
    \begin{center}
    \caption{{\bf Quantitative Evaluations.} We evaluate the appearance (PSNR, SSIM, LPIPS, FVD) and motion (EPE) of generated videos on the validation set of the DAVIS dataset. Please note that each method is trained from a different base model. 
    }
    
    \label{tbl:davis_eval_abridged}
    \setlength\tabcolsep{3pt}
    \resizebox{0.8\linewidth}{!}{
    \begin{tabular}{llccccc}
    \toprule
     
    \# Tracks & Method & PSNR $\uparrow$ & SSIM $\uparrow$ & LPIPS $\downarrow$ & FVD $\downarrow$ & EPE $\downarrow$ \\
    \midrule

    \multirow{ 3}{*}{N = 1} & Image Conductor & 11.468 & 0.145 & 0.529 & 1919.8 & 19.224  \\
    & DragAnything & 14.589 & 0.241 & 0.420 & 1544.9 & \textbf{9.135}  \\
    & Ours & \textbf{15.431} & \textbf{0.266} & \textbf{0.368} & \textbf{1445.2} & 14.619 \\
    \cmidrule(lr){1-7}

    \multirow{ 3}{*}{N = 16} & Image Conductor & 12.184 & 0.175 & 0.502 & 1838.9 & 24.263  \\
    & DragAnything & 15.119 & 0.305 & 0.378 & \textbf{1282.8} & 9.800  \\
    & Ours & \textbf{16.618} & \textbf{0.405} & \textbf{0.319} & 1322.0 & \textbf{8.319}  \\
    \cmidrule(lr){1-7}

    \multirow{ 3}{*}{N = 512} & Image Conductor & 11.902 & 0.132 & 0.524 & 1966.3 & 30.734  \\
    & DragAnything & 15.055 & 0.289 & 0.381 & 1379.8 & 10.948  \\
    & Ours & \textbf{18.968} & \textbf{0.583} & \textbf{0.229} & \textbf{688.7} & \textbf{4.055}  \\
    \cmidrule(lr){1-7}
    
    \multirow{ 3}{*}{N = 2048} & Image Conductor & 11.609 & 0.120 & 0.538 & 1890.7 & 33.561 \\
    & DragAnything & 14.845 & 0.286 & 0.397 & 1468.4 & 12.485  \\
    & Ours & \textbf{19.327} & \textbf{0.608} & \textbf{0.227} & \textbf{655.9} & \textbf{3.887}  \\
    
    \bottomrule
    \end{tabular}
   }
    \end{center}
\aftertab    
\end{table}

As shown in \cref{tbl:davis_eval_abridged}, our model outperforms the baselines in almost all cases.
On some examples, DragAnything performs better in terms of EPE with fewer tracks.
This is because DragAnything includes a module that effectively warps latents.
While this warping effect may result in accurate motion, it also creates visual artifacts as evidenced by the underperforming PSNR, SSIM, LPIPS, and FVD results. \supparxiv{Please refer to the \supp{} to see this effect. }{}We also provide numbers for 4 and 64 tracks in \cref{apdx:quantitative}, which we omit here for brevity.

\begin{table}[ht]
    \begin{center}
    \caption{{\bf Human Study.} We present \% win rates of our method against baselines in 2AFC human study results. Sample sizes are $N=103$, $N=103$, and $N=115$ for each column respectively.}
    
    \label{tbl:human_study}
    \setlength\tabcolsep{3pt}
    \resizebox{0.8\linewidth}{!}{
    \begin{tabular}{lccc}
    \toprule
     
    Method & Motion Adherence & Motion Quality & Visual Quality \\
    \midrule

    Image Conductor & 74.3\ci{1.1} & 80.5\ci{1.0} & 77.3\ci{1.0} \\
    Drag Anything & 74.5\ci{1.1} & 75.7\ci{1.1} & 73.7\ci{1.0} \\
    
    \bottomrule
    \end{tabular}
   }
    \end{center}
\aftertab
\end{table}

\subsection{Human Study}

We run a human study where we manually create a set of 30 inputs consisting of a single trajectory.
We run a two alternative forced choice test where we ask (1) which video follows the motion conditioning better (2) which video has more realistic motion and (3) which video has higher visual quality. There are 180 questions total, and win rates for our method as well as 95\% confidence intervals are presented in \cref{tbl:human_study}.
When considering both motion and appearance together, our approach is preferred over baselines in all categories. Implementation details can be found in \cref{apdx:implementation}.

\subsection{Ablations}

In \cref{tbl:ablations} we present an ablation where we train our model on only {\it Sparse} point trajectories (1-8 tracks) and {\it Dense + Sparse}, where the number of tracks is sampled logarithmically from $2^0$ to $2^{13}$. 
We find that dense training is most effective, especially for large number of tracks. Surprisingly, dense training is also better for sparse tracks. We hypothesize that this is because using sparse tracks gives so little training signal that it is more efficient to train on dense tracks, which then generalizes to sparser tracks, though this may be influenced by our usage of ControlNet and zero convolutions. We use a subset of the DAVIS evaluation from \cref{sec:davis_eval}, but we note that the numbers do not match as we use less data and fewer training steps for the ablations.

\begin{table}[t]
    \begin{center}
    \caption{{\bf Ablation.} We ablate the density of tracks during training and find that training on dense tracks works best for our model.
    }
    
    \label{tbl:ablations}
    \setlength\tabcolsep{3pt}
    \resizebox{0.82\linewidth}{!}{
    \begin{tabular}{llccccc}
    \toprule
     
    \# Tracks & Method & PSNR $\uparrow$ & SSIM $\uparrow$ & LPIPS $\downarrow$ & FVD $\downarrow$ & EPE $\downarrow$ \\
    \midrule

    \multirow{ 3}{*}{N = 4} & Sparse & 15.075 & 0.241 & 0.384 & \textbf{1209.2} & 30.712 \\
    & Dense + Sparse & 15.162 & 0.252 & 0.379 & 1230.6 & 29.466 \\
    & Dense & \textbf{15.638} & \textbf{0.296} & \textbf{0.349} & 1254.9 & \textbf{24.553} \\
    \cmidrule(lr){1-7}
    
    \multirow{ 3}{*}{N = 2048} & Sparse & 15.697 & 0.284 & 0.355 & 1322.0 & 26.724 \\
    & Dense + Sparse & 15.294 & 0.246 & 0.375 & 1267.8 & 27.931 \\
    & Dense & \textbf{19.197} & \textbf{0.582} & \textbf{0.230} & \textbf{729.0} & \textbf{4.806} \\
    
    \bottomrule
    \end{tabular}
   }
    \end{center}
\aftertab    
\end{table}

\section{Conclusion}

We have introduced a framework for motion-conditioned video generation that leverages flexible motion prompts -- spatio-temporal trajectories that can encode arbitrary motion complexity. Unlike prior work, this representation allows specifying sparse or dense motion for cameras, objects, or full scenes. We also introduce motion prompt expansion to translate high-level motion requests into detailed prompts. Our versatile approach enables applications like motion control, motion transfer, image editing, and showcasing emergent behaviors like realistic physics with a single unified model. Quantitative and human evaluation demonstrate the effectiveness of our framework.

\mypar{Acknowledgements} We would like to thank Sarah Rumbley, Roni Paiss, Jeong Joon Park, Liyue Shen, Stella Yu, Alyosha Efros, Boqing Gong, Daniel Watson, David Fleet, and Bill Freeman for their invaluable feedback and discussions.

{
    \small
    \bibliographystyle{ieeenat_fullname}
    \bibliography{main}
}

\newpage

\clearpage
\appendix
\setcounter{page}{1}
\maketitlesupplementary
\setcounter{figure}{0}
\renewcommand{\thefigure}{A\arabic{figure}}
\setcounter{table}{0}
\renewcommand{\thetable}{A\arabic{table}}

\supparxiv{Thank you for reading the supplementary material. \textbf{We highly encourage the reviewers to view sampled videos in the included website,} as many of our results cannot be fully appreciated as a sequence of static frames on paper. }{}

\section{Implementation Details}
\label{apdx:implementation}

\subsection{Architecture and Training}

We train our model for 70,000 steps using Adafactor~\cite{shazeer2018adafactor} with a learning rate of $1 \times 10^{-4}$. We do not use any learning rate decay. For the ControlNet, we copy Lumiere's encoder stack, add in zero convolutions as in~\cite{zhang2023adding}, and replace the first convolutional layer with a new layer that accepts a $T\times H \times W \times C$ conditioning signal. From the constraints of the base architecture, we have $T=80$ and $H = W = 128$. We set $C=64$. During training, we sample the number of input tracks uniformly from 1000 to 2000 inclusive. For each track we randomly assign a sinusoidal positional encoding~\cite{vaswani2017attention}, of 64 dimensions, by sampling integers without replacement from 0 to 16384 -- the maximum number of tracks for a $128 \times 128$ image, and using the corresponding positional embedding for that integer. Note that the encoding is completely randomly assigned. In particular, its spatial location has no bearing on its embedding.

All sampled videos are passed through Lumiere's spatial super resolution (SSR) model, resulting in a $1024 \times 1024$, 80 frame video at 16 frame per second, for a total of 5 seconds. We use Lumiere's SSR model as is, without finetuning it for motion conditioning, as we find that the $128 \times 128$ output of the base model already contains all of the motion conditioned dynamics.

\mypar{Data. } We train on an internal dataset of 2.2 million videos. We precompute trajectories on this dataset by center cropping each video to a square, resizing it to $256\times 256$, and then running BootsTAP~\cite{doersch2024bootstap} with a dense gird of query points, resulting in 16,384 tracks per video.  During training, a video is sampled, and then tracks are randomly sampled from this dataset.  During the Lumiere fine-tuning, videos are resized to match Lumiere's $128 \times 128$ input and output size.

\begin{table*}[t]
    \begin{center}
    \caption{{\bf Figure Details.} We provide details about qualitative samples shown in our figures, including text prompts fed to the model and licensing information. In general, these are sorted by the order that they appear in the paper, moving from left to right, top to bottom.}
    
    \label{tbl:figure_details}
    \setlength\tabcolsep{3pt}
    \resizebox{\linewidth}{!}{
    \begin{tabular}{@{\hspace{3mm}}lllllll}
    \toprule
     
    Description & Figure & Text Prompt & Source & URL & License & License URL \\
    \midrule

    two elephants & \cref{fig:teaser}, \cref{fig:depth} & \texttt{elephants} & Unsplash & \href{https://unsplash.com/photos/two-elephants-near-trees-XWTNFVCTS8E}{link} & Unsplash & \href{https://unsplash.com/license}{license} \\ \cmidrule(lr){1-7}
    
    owl & \cref{fig:teaser} & \texttt{a close up of a great horned owl} & Unsplash & \href{https://unsplash.com/photos/brown-owl-on-a-dark-place-BMO1SzQHWRs}{link} & Unsplash & \href{https://unsplash.com/license}{license} \\ \cmidrule(lr){1-7}
    
    brown bear & \cref{fig:teaser} & \texttt{a brown bear} & Unsplash & \href{https://unsplash.com/photos/brown-bear-near-grass-field-kZ8dyUT0h30}{link} & Unsplash & \href{https://unsplash.com/license}{license} \\ \cmidrule(lr){1-7}
    
    squirrel & \cref{fig:teaser} & \texttt{a squirrel sitting on the ground in the woods} & Unsplash & \href{https://unsplash.com/photos/a-squirrel-sitting-on-the-ground-in-the-woods-c_KfK8v9aQ4}{link} & Unsplash & \href{https://unsplash.com/license}{license} \\ \cmidrule(lr){1-7}
    
    golden retriever & \cref{fig:teaser}, \cref{fig:compositions} & \texttt{a golden retriever laying in the grass} & Unsplash & \href{https://unsplash.com/photos/adult-golden-retriever-sitting-on-green-grass-YI_iWr_12kE}{link} & Unsplash & \href{https://unsplash.com/license}{license} \\ \cmidrule(lr){1-7}
    
    man (motion source) & \cref{fig:teaser}, \cref{fig:motion_transfer} & \;-- & private correspondence & \;-- & permission granted & \;-- \\ \cmidrule(lr){1-7}
    
    macaque & \cref{fig:teaser}, \cref{fig:motion_transfer} & \texttt{a macaque monkey} & Unsplash & \href{https://unsplash.com/photos/shallow-focus-photography-of-monkey-ghD1Znf8gps}{link} & Unsplash & \href{https://unsplash.com/license}{license} \\ \cmidrule(lr){1-7}
    
    sand & \cref{fig:teaser}, \cref{fig:sparse} & \texttt{sand} & Unsplash & \href{https://unsplash.com/photos/focus-photo-of-brown-sand-eYWNaMffWHI}{link} & Unsplash & \href{https://unsplash.com/license}{license} \\ \cmidrule(lr){1-7}
    
    woman & \cref{fig:teaser}, \cref{fig:sparse} & \texttt{a woman} & private correspondence & \;-- & permission granted & \;-- \\ \cmidrule(lr){1-7}
    
    parrot & \cref{fig:sparse} & \texttt{a close up of a green parrot} & Unsplash & \href{https://unsplash.com/photos/a-close-up-of-a-green-parrot-with-a-red-beak-uhEwDYq0iM0}{link} & Unsplash & \href{https://unsplash.com/license}{license} \\ \cmidrule(lr){1-7}
    
    cow & \cref{fig:sparse}, \cref{fig:drag_editing}, \cref{fig:failures} & \makecell[l]{\texttt{a highland cow standing} \\ \texttt{in a grassy scottish wilderness}} & Unsplash & \href{https://unsplash.com/photos/a-long-haired-cow-standing-on-top-of-a-grass-covered-field-qiQmvXnQ_SE}{link} & Unsplash & \href{https://unsplash.com/license}{license} \\ \cmidrule(lr){1-7}
    
    skull & \cref{fig:drag_editing} & \texttt{a white skull on a black background} & Unsplash & \href{https://unsplash.com/photos/a-skeleton-is-standing-in-a-black-and-white-photo-B5Ddx7kx8yk}{link} & Unsplash & \href{https://unsplash.com/license}{license} \\ \cmidrule(lr){1-7}
    
    stool & \cref{fig:drag_editing} & \texttt{a living room} & Unsplash & \href{https://unsplash.com/photos/turned-off-flat-screen-television-on-white-dresser-dv9AoOYegRc}{link} & Unsplash & \href{https://unsplash.com/license}{license} \\ \cmidrule(lr){1-7}
    
    hot air balloons & \cref{fig:drag_editing} & \makecell[l]{\texttt{a serene scene of multiple hot air balloons} \\ \texttt{floating over Cappadocia, Turkey, during sunset}} & Unsplash & \href{https://unsplash.com/photos/a-bunch-of-hot-air-balloons-flying-in-the-sky-UeX_qw9lnzc}{link} & Unsplash & \href{https://unsplash.com/license}{license} \\ \cmidrule(lr){1-7}
    
    arches & \cref{fig:depth} & \makecell[l]{\texttt{arches in arches national park, with shrubbery} \\ \texttt{in the foreground and a bright blue sky}} & Unsplash & \href{https://unsplash.com/photos/arches-national-park-utah-during-daytime-Aydu-0d4Iwc}{link} & Unsplash & \href{https://unsplash.com/license}{license} \\ \cmidrule(lr){1-7}
    
    roses & \cref{fig:depth} & \texttt{a red rose} & Unsplash & \href{https://unsplash.com/photos/shallow-focus-photography-of-red-flower-LZCGRSQxn6E}{link} & Unsplash & \href{https://unsplash.com/license}{license} \\ \cmidrule(lr){1-7}
    
    cat & \cref{fig:primitives} & \texttt{a cat} & Unsplash & \href{https://unsplash.com/photos/white-and-gray-cat-IFxjDdqK_0U}{link} & Unsplash & \href{https://unsplash.com/license}{license} \\ \cmidrule(lr){1-7}
    
    frog & \cref{fig:primitives} & \texttt{a close up of a frog} & Unsplash & \href{https://unsplash.com/photos/brown-frog-in-close-up-photography-8GQ5ELnU-rM}{link} & Unsplash & \href{https://unsplash.com/license}{license} \\ \cmidrule(lr){1-7}
    
    horse & \cref{fig:compositions} & \texttt{a horse} & Unsplash & \href{https://unsplash.com/photos/shallow-focus-photography-of-black-donkey-vUpXnK5ufwg}{link} & Unsplash & \href{https://unsplash.com/license}{license} \\ \cmidrule(lr){1-7}
    
    Earth (motion source) & \cref{fig:motion_transfer} & \;-- & Pexels & \href{https://www.pexels.com/video/digital-animation-of-planet-earth-10880732/}{link} & Pexels & \href{https://www.pexels.com/license/}{license} \\ \cmidrule(lr){1-7}
    
    panda & \cref{fig:motion_transfer} & \texttt{a panda} & Unsplash & \href{https://unsplash.com/photos/a-panda-bear-in-the-grass-ScHhzUSG2x8}{link} & Unsplash & \href{https://unsplash.com/license}{license} \\ \cmidrule(lr){1-7}
    
    monkey (motion source) & \cref{fig:motion_transfer} & \;-- & Pexels & \href{https://www.pexels.com/video/close-up-footage-of-a-monkey-eating-its-food-7710018/}{link} & Pexels & \href{https://www.pexels.com/license/}{license} \\ \cmidrule(lr){1-7}
    
    trees & \cref{fig:motion_transfer} & \texttt{birds eye view of trees} & Unsplash & \href{https://unsplash.com/photos/aerial-view-of-green-trees-EI0pK6euSKE}{link} & Unsplash & \href{https://unsplash.com/license}{license} \\ \cmidrule(lr){1-7}
    
    chess & \cref{fig:failures} & \makecell[l]{\texttt{close-up of a chessboard with strong} \\ \texttt{depth of field. The white king} \\ \texttt{piece is in focus, surrounded by black pawns}} & Unsplash & \href{https://unsplash.com/photos/a-close-up-of-a-chess-board-with-pieces-on-it-y5nGbO1u8mA}{link} & Unsplash & \href{https://unsplash.com/license}{license} \\
    
    \bottomrule
    \end{tabular}
   }
    \end{center}
\aftertab
\end{table*}

\subsection{Qualitative Results}

We provide additional details for qualitative results in \cref{tbl:figure_details}, including text prompt and licensing information. All images and videos are used with permission and under open and free licenses. In addition, as can be seen we construct text prompts to describe the scene but not the motion, in order to limit the influence of text conditioning on the motion as much as possible.

\subsection{Mouse GUI}
\label{apdx:gui}

\begin{figure}[t]
    \centering
    \includegraphics[width=0.9\linewidth]{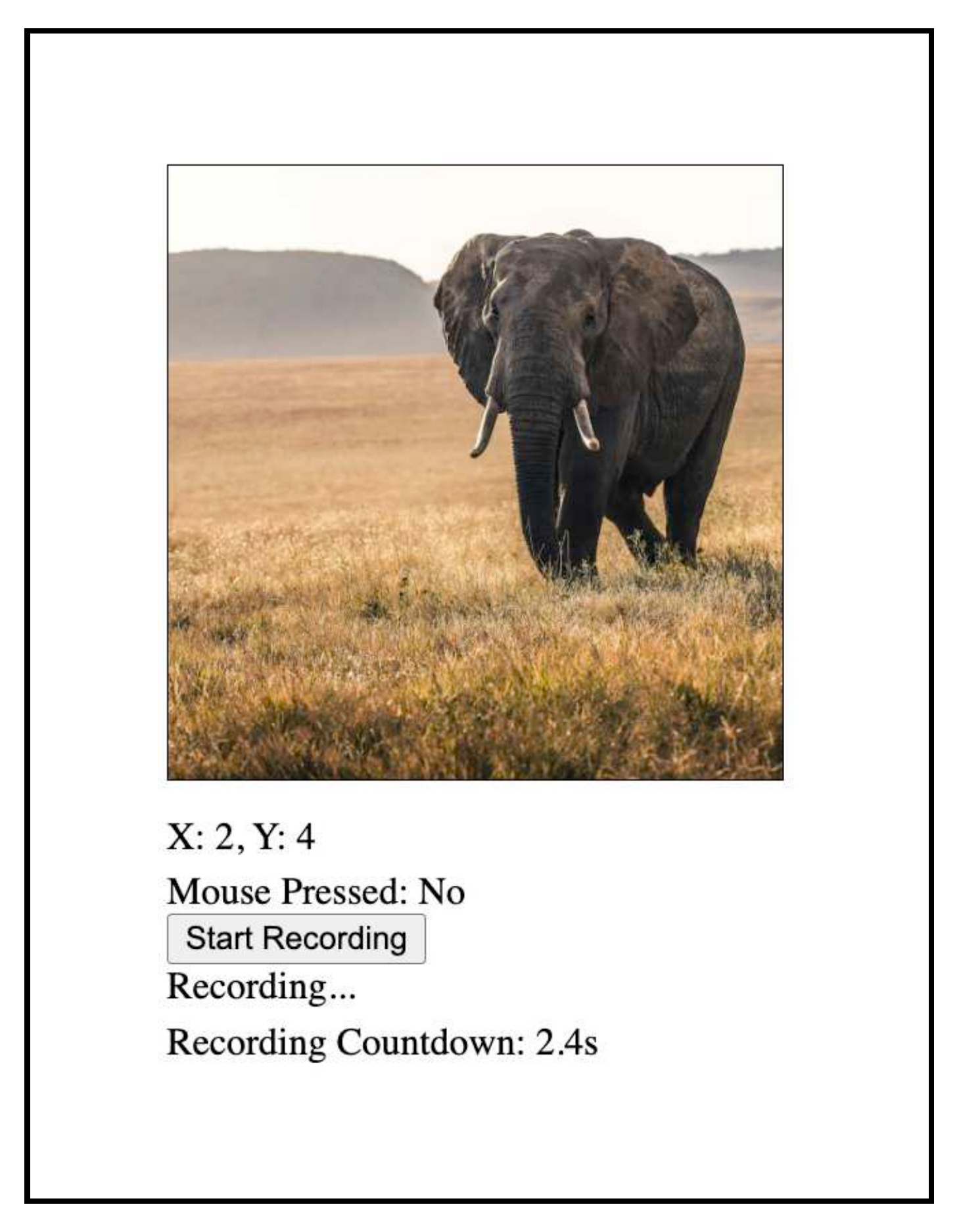}
    \caption{{\bf Mouse Motion GUI. } We show a screenshot of the GUI that we use to record mouse motions. For more information please see \cref{apdx:gui}. }
\label{fig:gui}
\afterfig
\end{figure}

We record mouse motions through a simple HTML GUI, which is shown in \cref{fig:gui}. It consists of a {\tt canvas} element which displays the first frame conditioning, labels that indicate the position of the mouse in the canvas, and whether or not it is currently being clicked, a button to start the recording, a countdown timer which gives three seconds before recording starts, and a second countdown timer which shows when the recording will end. We record 80 frames of mouse input to match the five seconds of video that our model outputs at 16 frames per second. For each frame we record the mouse $(x, y)$ position, and a flag indicating whether the mouse is being clicked.

\subsection{Interacting with and Drag Editing Images}

In order to feed mouse motions to our model, we create a grid of tracks that is centered on the mouse whenever it is being dragged. The user may choose the stride of these tracks, and the size of the grid. We use a square grid of tracks for simplicity. In addition, a user may choose to have the tracks ``persist," in that before and after the mouse drag the tracks remain. This is useful in cases where objects should stay in place after a drag. A user may also place down a grid of tracks to ``pin" the background in place. Note that this setup is identical to how we obtain the ``drag-based image editing" results.

\subsection{Geometric Primitives}

To make spherical tracks we take points on a sphere and follow them as the sphere is spun. This gives us a trajectory of 3D points, which when orthographically projected gives us 2D tracks. The density of the points, the radius of the sphere, and the location of the sphere are determined by the user. Mouse motions are converted to sphere spins by rotating the sphere through a single axis such that the initial mouse location matches with the current mouse location at each frame. This uniquely defines a rotation and ensures that the sphere tracks the mouse.

\subsection{Camera Control}
In order to obtain camera control, we run a monocular depth estimator on the first frame input to the model. This gives us camera intrinsics as well as depths, allowing us to un-project into a point cloud. We then project this point cloud onto a sequence of camera poses forming the desired camera trajectory, resulting in 2D point tracks. In addition, we run z-buffering to determine occlusions, where only the closest point that has been projected to some neighborhood is visible while all other points in that neighborhood are occluded---unless that point is sufficiently close to the visible point. This requires choosing a radius for the neighborhood size, and a threshold for a point to remain visible if it is close enough to the visible point. Both are set manually to constant values that we find to work well for all examples.

We also discuss translating mouse motion to camera motion. This is done by having the camera move in such a way that the mouse is always above the same point. Because this is underdetermined, we also add the constraint that the camera should stay fixed in the vertical plane. Note that this is not the only constraint possible. Other constraints may restrict the camera to the surface of a sphere around the scene for example.

\subsection{Track Sparsity}

For camera control and motion transfer motion prompts, we obtain a dense set of tracks. Empirically, we find that it is helpful to randomly subsample these tracks, as using too many tracks suppresses the video model's learned priors from working, while using too few affords too little control. Somewhere in the middle is a sweet spot. For example, for the majority of the depth-based motion prompts, we use 1024 tracks, which we find offers a good balance between control and emergent video prior effects. In other cases, such as transferring the motion of the person's face in \cref{fig:motion_transfer}, we find that fewer tracks is helpful in dealing with misalignments between the source video and the input first frame. Finally, for out-of-domain motion transfer as in the {\it monkey chewing} example in \cref{fig:motion_transfer}, we find that very dense tracks help. We use 1500 tracks, as we need a lot of control to apply such an unnatural motion to the first frames.

\subsection{Davis Eval}
We conduct a quantitative evaluation of first frame, text, and track conditioned video generation using the DAVIS validation dataset, with a subset of results in \cref{tbl:davis_eval_abridged} and full results in \cref{tbl:davis_eval_full}. The validation dataset contains 30 videos from a wide range of scenes, involving subjects from sports to humans to animals to cars. In order to create inputs for the models, we extract tracks using BootsTAP~\cite{doersch2024bootstap}. First frame inputs are square crops of the first frames of the videos, and text prompts are the titles of the videos given by the dataset and typically consist of a word or two. For each evaluation for a given number of tracks, we randomly sample that number of tracks for conditioning.

To ensure a fair comparison, we make the following accommodations for baselines. In addition to a first frame, tracks, and a text prompt, DragAnything requires segmentation masks for objects that the tracks move. To get this, we use the provided ground truth segmentations in the DAVIS dataset. Image Conductor is a finetuned version of AnimateDiff and is trained on videos of resolution $384 \times 256$. We initially gave the model $256 \times 256$ images, and found that we got reasonable results. However, we experimented with reflection padding the input frame to $384 \times 256$ and cropping the output, which gave slightly better results which we report. 

\begin{figure}[t]
    \centering
    \includegraphics[width=\linewidth]{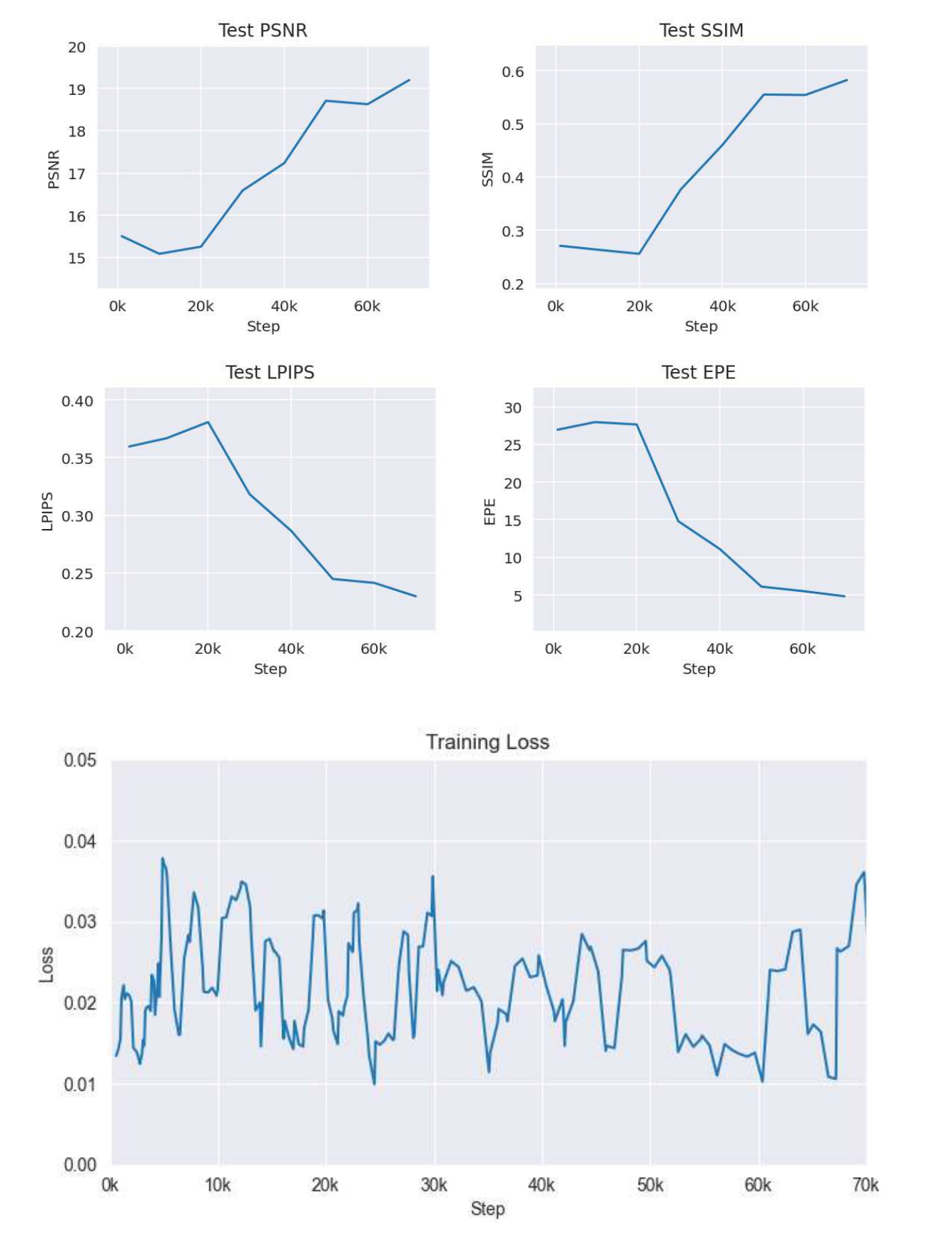}
    \caption{{\bf Test and Train Metrics. } Here we plot out training loss, along with PSNR, SSIM, LPIPS, and EPE on our DAVIS test set. Note that there is no correlation between the training loss and the test metrics, and that the test metrics show no signs of improvement until step 20,000 at which point the network learns quite rapidly.}
\label{fig:losses}
\afterfig
\end{figure}

\subsection{Human Studies}

To perform the human studies, we handcraft 30 inputs with diverse image subjects and input motions. Motions consist of a single uninterrupted trajectory. Text prompts are designed to describe the image, but not the desired motion, to factor out the influence of text as much as possible. DragAnything requires masks, which we obtain by running SAM~\cite{kirillov2023segany} on the first frame with the initial location of the tracks as query points. For our method, we turn the trajectory into a grid of tracks as described above. We then feed these inputs to the models and take a single random sample. We follow the same protocol as above for Image Conductor. This results in 30 samples for each model and 90 samples in total. 

We run a two alternative forced choice (2AFC) test between our model and the baselines. We display a sample from our method and a sample from the baseline in a random order with a video of the corresponding motion conditioning in the middle, visualized as a moving red dot. Participants are then asked three questions. Verbatim, we ask (1) {\it Which video better follows the motion of the red dot?} (2) {\it Which video has the more realistic motion?} (3) {\it Which video is of higher visual quality?} These questions are designed to measure the adherence of the motion to the conditioning, the quality of the motion, and the overall visual quality of video, respectively. This results in a total of 180 questions.

We recruit participants for our study through Amazon Mechanical Turk (MTurk). To ensure responses are of high quality, we add three ``vigilance" questions with clearly correct answers. We discard all responses that fail any of these three questions. Each question is conducted as a separate study, and the resulting number of participants are $N=103$, $N=103$, and $N=115$ for each question respectively. This results in a total of 19,260 answers.

\section{Training Observations}
\label{apdx:training}

In training, we observe similar behavior as noted in ControlNet~\cite{zhang2023adding} and ControlNext~\cite{peng2024controlnext}: 1) training loss does not directly correlate with model performance, and 2) ``sudden convergence'' where in a few epochs the model goes from not adhering to control signal to full adherence. ControlNext identifies both of these behaviors as coming from the zero initialization and offers cross normalization as a potential solution. We believe this and other future control scheme is a promising avenue for future work in track conditioned video generation. We show training loss and test metrics in \cref{fig:losses}. As can be seen, the training loss is fairly inscrutable, while the test losses do not begin to decrease until step 20,000.

\section{Full Quantitative Results}
\label{apdx:quantitative}

In \cref{sec:quantitative} we present DAVIS evaluation results for $N = \{1, 16, 512, 2048\}$. In \cref{tbl:davis_eval_full} we present results for $ N = 4 $ and $ N = 64 $ as well, which we omit from the main paper for brevity.

\begin{table}[ht]
    \begin{center}
    \caption{{\bf Quantitative Evaluations.} We evaluate the appearance (PSNR, SSIM, LPIPS, FVD) and motion (EPE) of generated videos using the validation set of the DAVIS dataset. Please note that each method is trained from a different base model.
    }
    
    \label{tbl:davis_eval_full}
    \setlength\tabcolsep{3pt}
    \resizebox{\linewidth}{!}{
    \begin{tabular}{llccccc}
    \toprule
     
    \# Tracks & Method & PSNR $\uparrow$ & SSIM $\uparrow$ & LPIPS $\downarrow$ & FVD $\downarrow$ & EPE $\downarrow$ \\
    \midrule

    \multirow{ 3}{*}{N = 1} & Image Conductor & 11.468 & 0.145 & 0.529 & 1919.8 & 19.224  \\
    & DragAnything & 14.589 & 0.241 & 0.420 & 1544.9 & \textbf{9.135}  \\
    & Ours & \textbf{15.431} & \textbf{0.266} & \textbf{0.368} & \textbf{1445.2} & 14.619 \\
    \cmidrule(lr){1-7}
    
    \multirow{ 3}{*}{N = 4} & Image Conductor & 12.017 & 0.176 & 0.499 & 1735.0 & 18.921 \\
    & DragAnything & 15.040 & 0.272 & 0.397 & 1497.2 & \textbf{8.946}  \\
    & Ours & \textbf{15.820} & \textbf{0.319} & \textbf{0.353} & \textbf{1207.7} & 12.985 \\
    \cmidrule(lr){1-7}
    
    \multirow{ 3}{*}{N = 16} & Image Conductor & 12.184 & 0.175 & 0.502 & 1838.9 & 24.263  \\
    & DragAnything & 15.119 & 0.305 & 0.378 & \textbf{1282.8} & 9.800  \\
    & Ours & \textbf{16.618} & \textbf{0.405} & \textbf{0.319} & 1322.0 & \textbf{8.319}  \\
    \cmidrule(lr){1-7}
    
    \multirow{ 3}{*}{N = 64} & Image Conductor & 12.513 & 0.180 & 0.503 & 1947.7 & 26.316  \\
    & DragAnything & 14.499 & 0.274 & 0.393 & 1342.0 & 10.642  \\
    & Ours & \textbf{18.000} & \textbf{0.513} & \textbf{0.265} & \textbf{951.4} & \textbf{4.127}  \\
    \cmidrule(lr){1-7}
    
    \multirow{ 3}{*}{N = 512} & Image Conductor & 11.902 & 0.132 & 0.524 & 1966.3 & 30.734  \\
    & DragAnything & 15.055 & 0.289 & 0.381 & 1379.8 & 10.948  \\
    & Ours & \textbf{18.968} & \textbf{0.583} & \textbf{0.229} & \textbf{688.7} & \textbf{4.055}  \\
    \cmidrule(lr){1-7}
    
    \multirow{ 3}{*}{N = 2048} & Image Conductor & 11.609 & 0.120 & 0.538 & 1890.7 & 33.561 \\
    & DragAnything & 14.845 & 0.286 & 0.397 & 1468.4 & 12.485  \\
    & Ours & \textbf{19.327} & \textbf{0.608} & \textbf{0.227} & \textbf{655.9} & \textbf{3.887}  \\
    
    \bottomrule
    \end{tabular}
   }
    \end{center}
\aftertab    
\end{table}

\section{Human Pose Control}
\label{apdx:human_pose}

\begin{figure}[t]
    \centering
\newcommand{\datasetimgwidth}{0.185\linewidth}
\begin{tabular}{c @{\hskip 0.2em} c @{\hskip 0.2em} c @{\hskip 0.2em} c@{\hskip 0.2em} c}
	\includegraphics[width=\datasetimgwidth]{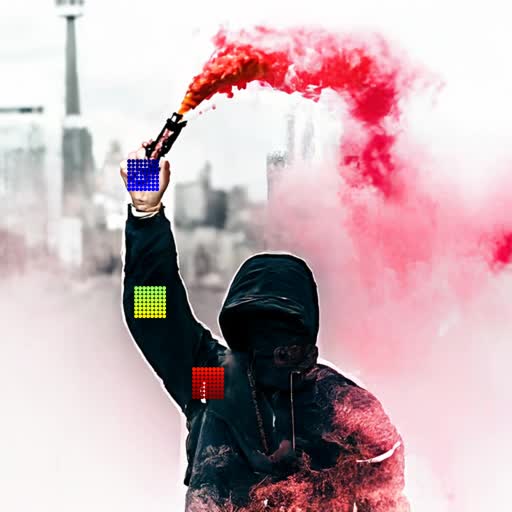} & 
	\includegraphics[width=\datasetimgwidth]{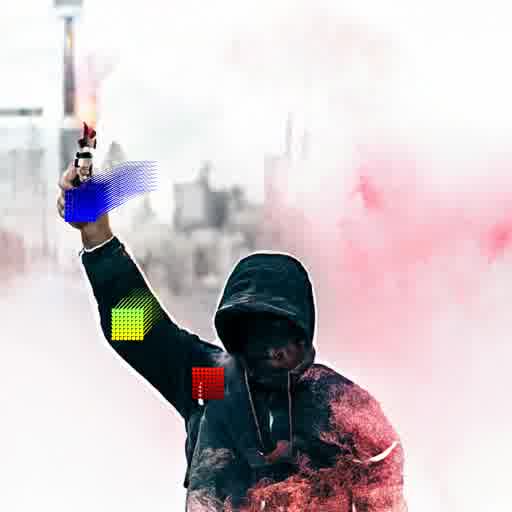} & 
	\includegraphics[width=\datasetimgwidth]{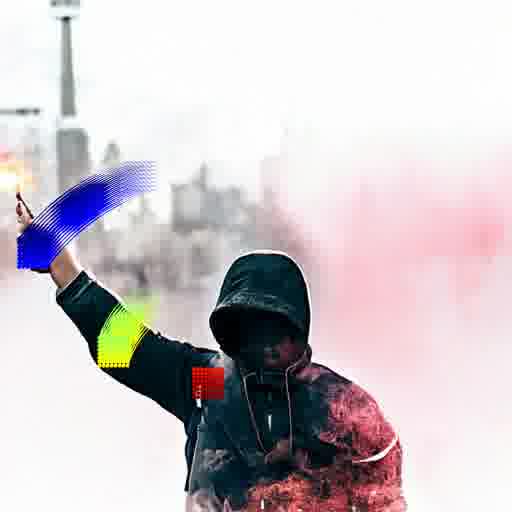} & 
	\includegraphics[width=\datasetimgwidth]{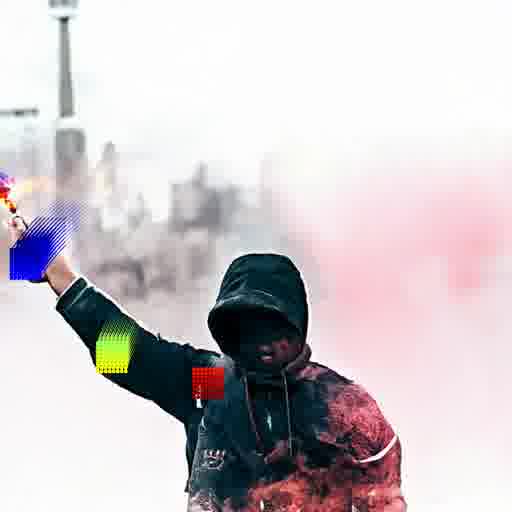} & 
	\includegraphics[width=\datasetimgwidth]{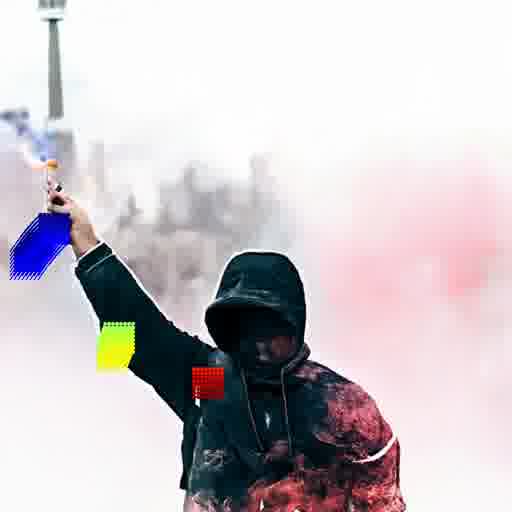} \\
    
    \includegraphics[width=\datasetimgwidth]{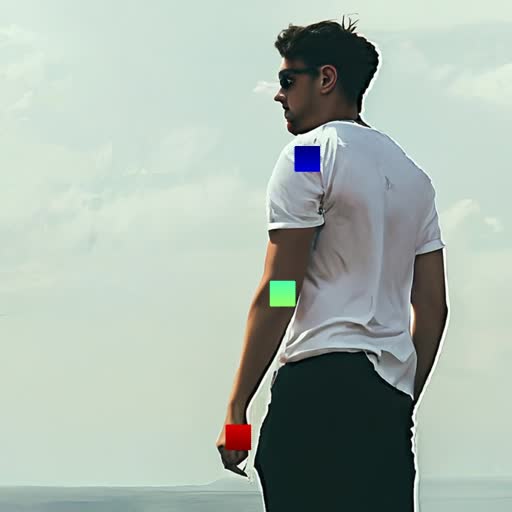} & 
	\includegraphics[width=\datasetimgwidth]{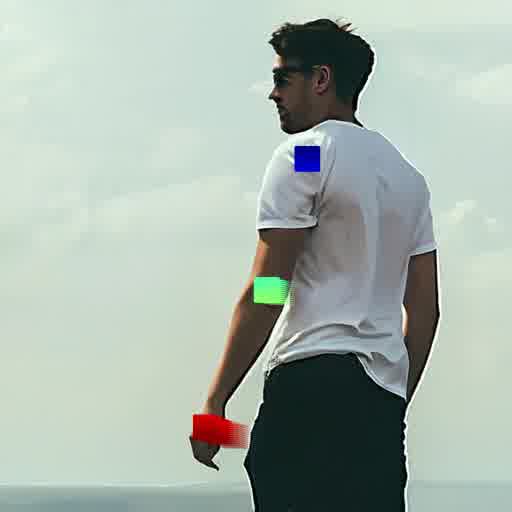} & 
	\includegraphics[width=\datasetimgwidth]{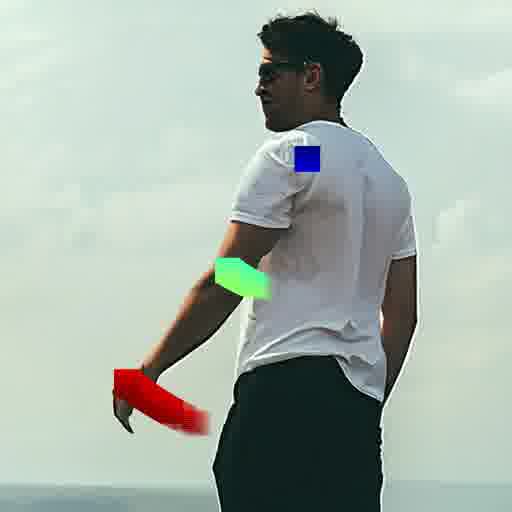} & 
	\includegraphics[width=\datasetimgwidth]{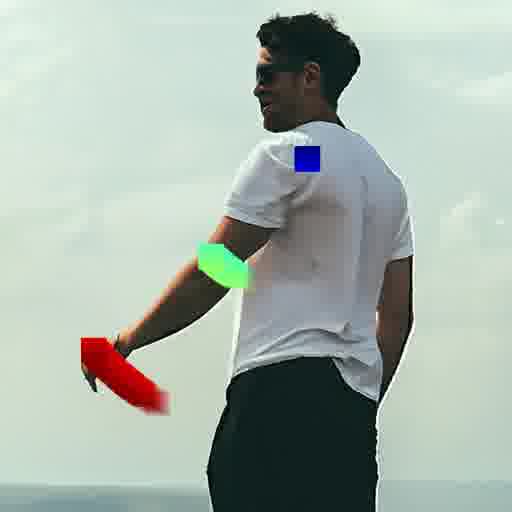} & 
	\includegraphics[width=\datasetimgwidth]{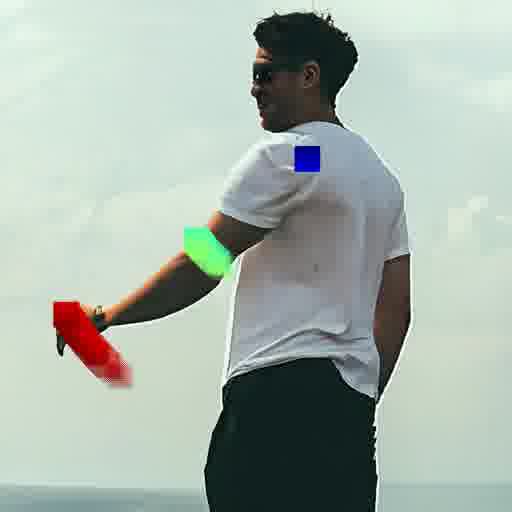} \\
    
	\includegraphics[width=\datasetimgwidth]{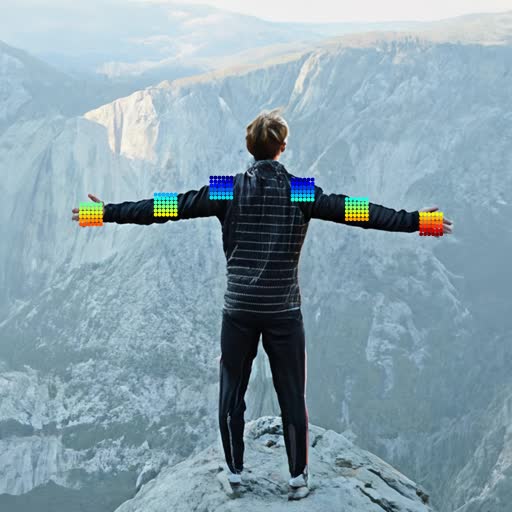} & 
	\includegraphics[width=\datasetimgwidth]{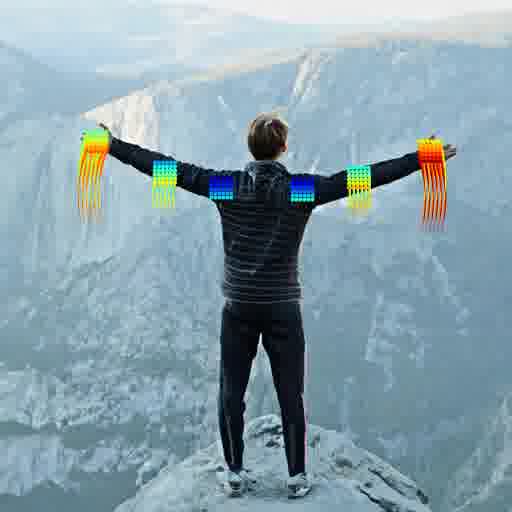} & 
	\includegraphics[width=\datasetimgwidth]{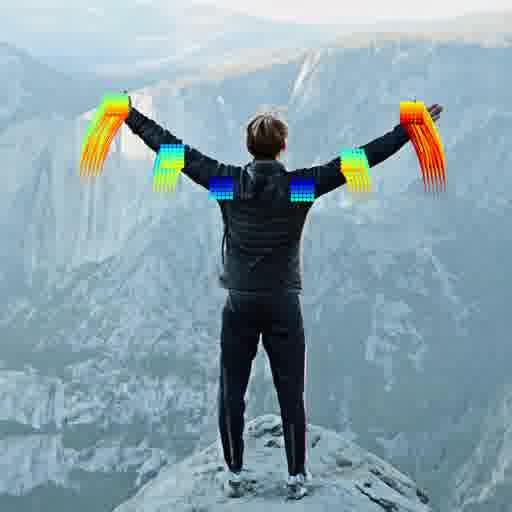} & 
	\includegraphics[width=\datasetimgwidth]{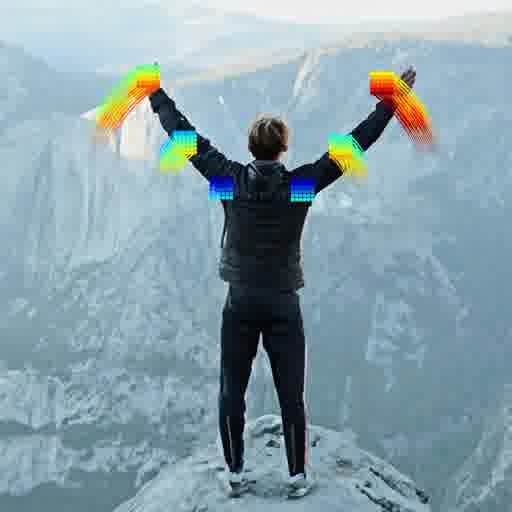} & 
	\includegraphics[width=\datasetimgwidth]{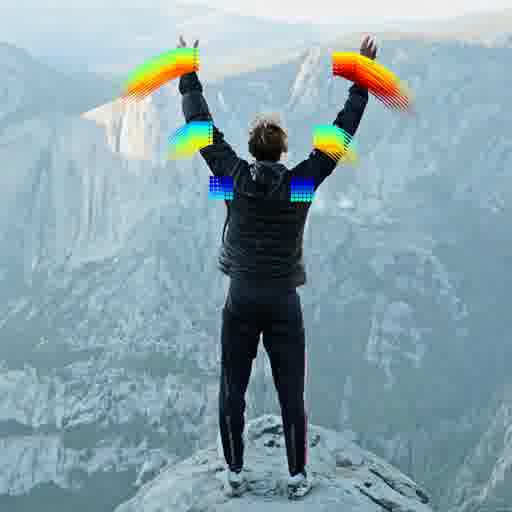} \\

\end{tabular}

    \caption{{\bf Pose Conditioning.} We estimate human pose, animate it, translate it to tracks, and then feed it to our model. In each row, we show frames from generated videos with input tracks overlaid on top. }
\label{fig:pose}
\afterfig
\end{figure}

We show results on using our method to control humans through human pose estimated keypoints in \cref{fig:pose}. To do this, we first estimate the pose with an off-the-shelf model, and then apply motions to desired keypoints and feed it to our model.

\section{Motion Magnification}
\label{apdx:motionmag}

\begin{figure}[h]
    \centering
    \includegraphics[width=\linewidth]{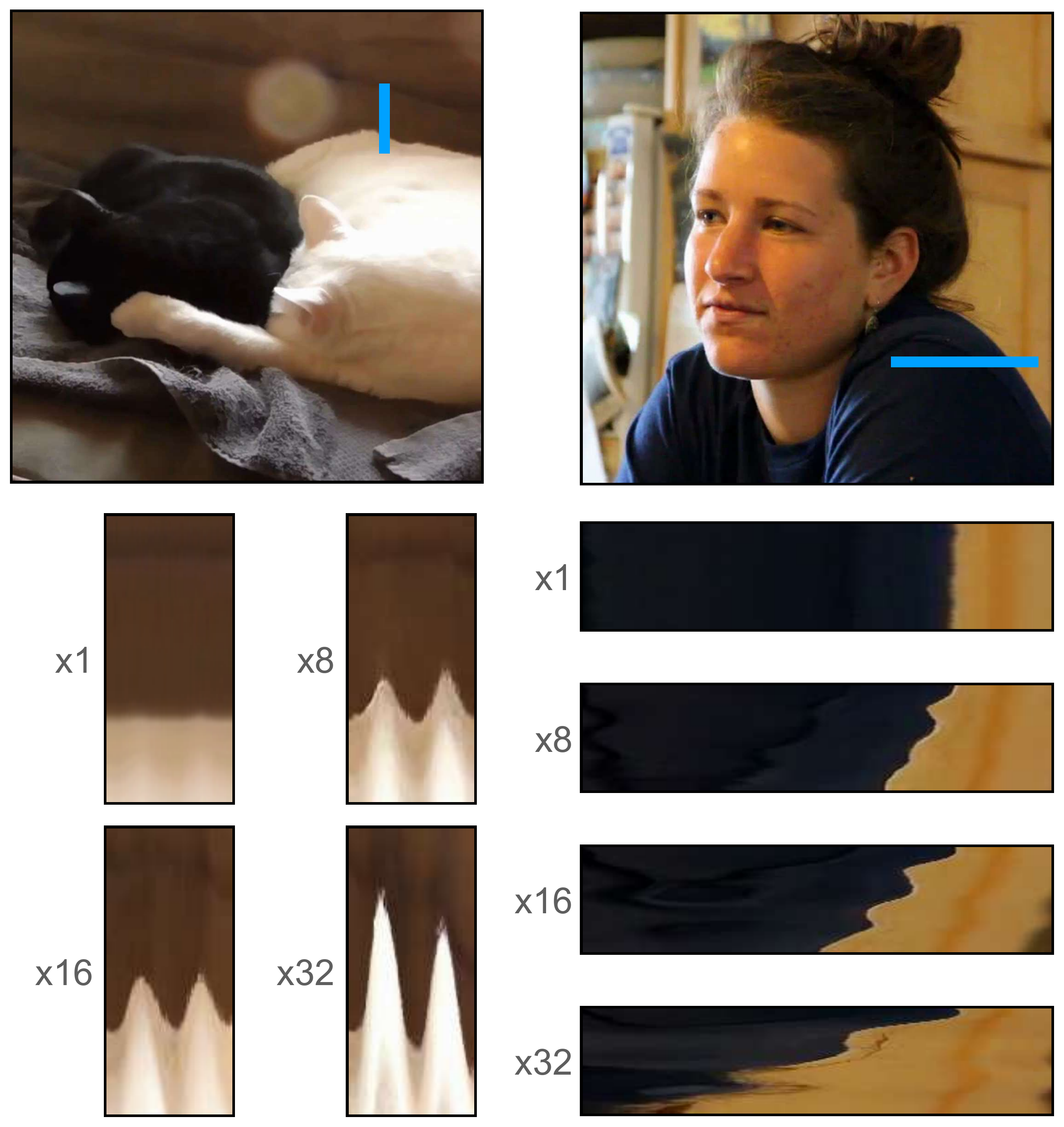}
    \caption{{\bf Motion Magnification. } We show the result of using our model to perform motion magnification. We show the first frame of two videos, and space-time slices through the blue line at different magnification factors.}
\label{fig:motionmag}
\afterfig
\end{figure}

One additional application of our model is motion magnification~\cite{liu2005motion,wu2012eulerian,wadhwa2013phase,oh2018learning,pan2023selfsupervised}. This task involves taking a video with subtle motions and generating a new video in which these motions have been magnified, so that they are easier to see. We do this by running a tracking algorithm~\cite{doersch2023tapir} on an input video, smoothing the tracks by applying a Gaussian blur over space and time, and then magnifying the resulting tracks. We then feed the first frame of the input video and the magnified tracks to our model. We show results, along with space-time slices, in \cref{fig:motionmag}. We found that smoothing was necessary to reduce noise in the estimated tracks. As a result the magnified tracks are not exactly at the specified magnification factor, but nonetheless are qualitatively useful in revealing subtle motions. We expect more accurate point tracking algorithms will remove the need for this smoothing step.

\end{document}